\documentclass[journal]{IEEEtran}
\usepackage[utf8]{inputenc}
\usepackage[T1]{fontenc}

\pdfoutput=1

\usepackage{amsmath}
\usepackage{mnick}
\usepackage{bbm}
\usepackage{tabularx,booktabs}
\usepackage{tablefootnote}
\usepackage{capt-of}
\usepackage[caption=false]{subfig}
\usepackage[pdftex]{graphicx}
\usepackage{mathtools}
\usepackage{microtype}
\makeatletter
\g@addto@macro{\UrlBreaks}{\UrlOrds}
\makeatother

\usepackage{tikz}
\usetikzlibrary{arrows,topaths,calc}
\usetikzlibrary{shadows,positioning}
\usetikzlibrary{shapes}
\usetikzlibrary{backgrounds}
\usetikzlibrary{decorations.pathreplacing}
\usepackage{tkz-graph}

\renewcommand{\loss}{\mathcal{L}}

\newcommand{\Nentities}{N_e}
\newcommand{\Nreln}{N_r}
\newcommand{\Nsamples}{N_d}
\newcommand{\dimentities}{H_e}
\newcommand{\dimreln}{H_r}
\newcommand{\dimA}{H_a}

\newcommand{\calD}{{\mathcal{D}}}

\newcommand{\mat}[1]{\mathbf{#1}}
\newcommand{\myvec}[1]{\mathbf{#1}}
\newcommand{\mymatrix}[1]{\mathbf{#1}}
\newcommand{\mytensor}[1]{\underline{\mathbf{#1}}}

\newcommand{\myvecsym}[1]{\bm{#1}}

\newcommand{\vone}{\myvecsym{1}}
\newcommand{\vtheta}{\myvecsym{\theta}}
\newcommand{\vphi}{\myvecsym{\phi}}

\newcommand{\va}{\myvec{a}}
\newcommand{\vb}{\myvec{b}}
\newcommand{\ve}{\myvec{e}}

\newcommand{\vg}{\myvec{g}}
\newcommand{\vh}{\myvec{h}}
\newcommand{\vr}{\myvec{r}}
\newcommand{\vu}{\myvec{u}}
\newcommand{\vv}{\myvec{v}}
\newcommand{\vw}{\myvec{w}}
\newcommand{\vx}{\myvec{x}}
\newcommand{\vy}{\myvec{y}}

\newcommand{\mA}{\mymatrix{A}}
\newcommand{\mB}{\mymatrix{B}}
\newcommand{\mC}{\mymatrix{C}}
\newcommand{\mE}{\mymatrix{E}}
\newcommand{\mF}{\mymatrix{F}}
\newcommand{\mI}{\mymatrix{I}}

\newcommand{\mW}{\mymatrix{W}}

\newcommand{\mY}{\mymatrix{Y}}

\newcommand{\tA}{\mytensor{A}}
\newcommand{\tB}{\mytensor{B}}
\newcommand{\tF}{\mytensor{F}}

\newcommand{\tY}{\mytensor{Y}}

\definecolor{ccls1}{RGB}{215,48,39} \definecolor{ccls2}{RGB}{252,141,89}
\definecolor{ccls3}{RGB}{254,224,144} \definecolor{ccls4}{RGB}{224,243,248}
\definecolor{ccls5}{RGB}{145,191,219} \definecolor{ccls6}{RGB}{69,117,180}

\definecolor{ccls41}{RGB}{215,25,28} \definecolor{ccls42}{RGB}{253,174,97}
\definecolor{ccls43}{RGB}{171,221,164}  \definecolor{ccls44}{RGB}{43,131,186}

\definecolor{ccls31}{RGB}{252,141,89} \definecolor{ccls32}{RGB}{255,255,191}
\definecolor{ccls33}{RGB}{153,213,148}

\usepackage{varwidth,array,ragged2e}
\newcolumntype{Y}{>{\centering\arraybackslash}X}
\newcolumntype{L}{>{\varwidth[c]{\linewidth}}l<{\endvarwidth}}

\usepackage{epigraph}
\usepackage{subfig}

\usepackage[numbers]{natbib}
\renewcommand\citet{\citep}

\crefname{secinapp}{appendix}{appendices}
\Crefname{secinapp}{Appendix}{Appendices}

\newcommand{\praleft}[1]{\ensuremath{\xleftarrow{\text{#1}}}}
\newcommand{\praright}[1]{\ensuremath{\xrightarrow{\text{#1}}}}
\newenvironment{triplelist}{\bgroup
  \itshape \begin{center}\begin{tabular}{lll}
 {\footnotesize \textbf{subject}} & {\footnotesize \textbf{predicate}} & {\footnotesize \textbf{object}}\\
  \cmidrule(lr){1-1} \cmidrule(lr){2-2} \cmidrule (lr){3-3}
}{\end{tabular}\end{center}\egroup}
\newenvironment{rulelist}{\bgroup
  \itshape \begin{center}\begin{tabular}{ll}}{\end{tabular}\end{center}\egroup}

\renewcommand{\mdef}{\ \coloneqq\ }

\begin{document}
%

\title{A Review of Relational Machine Learning\\
for Knowledge Graphs}

%

\author{Maximilian Nickel, Kevin Murphy, Volker Tresp,  Evgeniy Gabrilovich 
\thanks{Maximilian Nickel is with LCSL, Massachusetts Institute of Technology and Istituto Italiano di Tecnologia.}
\thanks{Volker Tresp is with Siemens AG, Corporate Technology and
    the Ludwig Maximilian University Munich.}
\thanks{Kevin Murphy and Evgeniy Gabrilovich are with Google Inc.}
\thanks{Manuscript received April 7, 2015; revised August 14, 2015.}}

\maketitle

\begin{abstract}
Relational machine learning studies methods for the statistical analysis
of relational, or graph-structured, data.
In this paper, we provide a
review of how such statistical models can be ``trained'' on
large knowledge graphs,
and then used to predict new facts about the world (which is
equivalent to predicting new edges in the graph).
In particular, we discuss two fundamentally different kinds of
statistical relational models, both of which can scale to  massive datasets.
The first is based on latent feature models such as tensor
  factorization and multiway neural networks.
The second is based on mining observable patterns in the graph.
We also show how to combine these latent and observable models to get improved
modeling power at decreased computational cost.
Finally, we discuss how such statistical models of graphs can be combined with text-based
information extraction methods for automatically constructing
knowledge graphs from the Web.
To this end, we also discuss Google's Knowledge Vault project as an example of such combination.
\end{abstract}




%
\IEEEpeerreviewmaketitle

\section{Introduction}
\label{sec:intro}

\begin{epigraphs}
  \qitem{%
    I am convinced that the crux of the problem of learning is
    recognizing relationships and being able to use them.
  }{\textit{Christopher Strachey in a letter to Alan Turing, 1954}}
\end{epigraphs}

%
%
%
%

\IEEEPARstart{T}{raditional} machine learning algorithms take as input a feature
vector, which represents an object in terms of numeric or categorical
attributes. The main learning task is to learn a mapping from this feature
vector to an output prediction of some form. This could be class labels, a
regression score, or an unsupervised cluster id or latent vector (embedding). In
Statistical Relational Learning (SRL), the representation of an object can
contain its relationships to other objects. Thus the data is in the form of a
{\em graph}, consisting of nodes (entities) and labelled edges (relationships
between entities). The main goals of SRL include prediction of missing edges,
prediction of properties of nodes, and clustering nodes based on their
connectivity patterns. These tasks arise in many settings such as analysis of
social networks and biological pathways. For further information on SRL see
\citet{getoor_introduction_2007,dzeroski_relational_2001,de_raedt_logical_2008}.

In this article, we review a variety of techniques from the SRL community and
explain how they can be applied to large-scale \emph{knowledge graphs} (KGs),
i.e., graph structured \emph{knowledge bases} (KBs) that store factual
information in form of relationships between entities. Recently, a large number
of knowledge graphs have been created, including
YAGO~\citep{suchanek_yago:_2007}, DBpedia~\citep{auer_dbpedia:_2007},
NELL~\citep{carlson_toward_2010}, Freebase~\citep{bollacker_freebase:_2008}, and
the Google Knowledge Graph~\citep{singhal_introducing_2012}. As we discuss in
Section~\ref{sec:kg}, these graphs contain millions of nodes and billions of
edges. This causes us to focus on {\em scalable} SRL techniques, which take time
that is (at most) linear in the size of the graph.

We can apply SRL methods to existing KGs to learn a model that can predict new
facts (edges) given existing facts. We can then combine this approach with
information extraction methods that extract ``noisy'' facts from the Web (see
e.g., \citep{weikum_information_2010,fan_akbc-wekex_2012}). For example, suppose
an information extraction method returns a fact claiming that Barack Obama was
born in Kenya, and suppose (for illustration purposes) that the true place of
birth of Obama was not already stored in the knowledge graph. An SRL model can
use related facts about Obama (such as his profession being US President) to
infer that this new fact is unlikely to be true and should be discarded. This
provides us a way to ``grow'' a KG automatically, as we explain in more detail
in Section~\ref{sec:kv}.

The remainder of this paper is structured as follows. In
\cref{sec:knowledge-graphs} we introduce knowledge graphs and some of their
properties. \Cref{sec:srl} discusses SRL and how it can be applied to knowledge
graphs. There are two main classes of SRL techniques: those that capture the
correlation between the nodes/edges using latent variables, and those that
capture the correlation directly using statistical models based on the
observable properties of the graph. We discuss these two families in
\cref{sec:lfm} and \cref{sec:observable}, respectively. \Cref{sec:combining}
describes methods for combining these two approaches, in order to get the best
of both worlds. \cref{sec:training} discusses how such models can be trained on
KGs. In \cref{sec:MRF} we discuss relational learning using Markov Random
Fields. In \cref{sec:exper} we describe how SRL can be used in automated
knowledge base construction projects. In \cref{sec:extension} we discuss
extensions of the presented methods, and \cref{sec:concl} presents our
conclusions.


\section{Knowledge Graphs}
\label{sec:knowledge-graphs}
\label{sec:kg}

In this section, we introduce knowledge graphs,
and discuss how they are represented, constructed, and used.

\subsection{Knowledge representation}
Knowledge graphs model information in the form of entities and relationships
between them. This kind of relational knowledge representation has a long
history in logic and artificial intelligence~\citep{davis_what_1993}, for
example, in semantic networks~\citep{sowa2006semantic} and
frames~\citep{minsky1975framework}. More recently, it has been used in the
Semantic Web community with the purpose of creating a ``web of data'' that is
readable by machines~\citep{berners-lee_semantic_2001}. While this vision of the
Semantic Web remains to be fully realized, parts of it have been achieved. In
particular, the concept of \emph{linked
  data}~\cite{berners-lee_linked_2006,bizer_linked_2009} has gained traction, as
it facilitates publishing and interlinking data on the Web in relational form
using the W3C Resource Description Framework
(RDF)~\cite{klyne_resource_2004,cyganiak_rdf_2014}. (For an introduction to
knowledge representation, see
e.g.~\citep{davis_what_1993,brachman2004knowledge,sowa1999knowledge}).

In this article, we will loosely follow the RDF standard and represent 
facts in the form of binary relationships, \mbox{in particular} \textit{{(subject, predicate, object)}} (SPO) triples, where \textit{subject} and \textit{object} are
entities and \textit{predicate} is the relation between them. (We discuss how to
represent higher-arity relations in Section~\ref{sec:higherOrderRelations}.) The
existence of a particular SPO triple indicates an existing fact, i.e., that the
respective entities are in a relationship of the given type. For instance, the
information
\begin{quote}
  \itshape Leonard Nimoy was an actor who played the character Spock in the
  science-fiction movie Star Trek
\end{quote}
can be expressed via the following set of SPO triples:
\begin{triplelist}
 (LeonardNimoy, & profession,& Actor) \\
 (LeonardNimoy, & starredIn, & StarTrek) \\
 (LeonardNimoy, & played, & Spock) \\
 (Spock, & characterIn, & StarTrek) \\
 (StarTrek, & genre, & ScienceFiction)
\end{triplelist}

\begin{figure}[tb]
  \centering
  \begin{tikzpicture}[scale=1,thick]
    \GraphInit[vstyle=Classic]
    \tikzset{vertex/.style =
      {draw=black,shape=circle,fill=white,minimum
        size=15pt,circular drop shadow,inner sep=0pt,
        prefix after command= {\pgfextra{\tikzset{every label/.style={font=\rmfamily\small}}}}
    }}
    \node [vertex,label=below:Leonard Nimoy] (nimoy) {};
    \node [vertex,label=above:Spock,above=1.5cm of nimoy] (spock) {};
    \node [vertex,label=below:Star Trek\vphantom{y},right=2cm of nimoy] (st) {};
    \node [vertex,label=above:Science Fiction\vphantom{y},above right=1.7cm and .5cm of st] (sf) {};

    \node [vertex,label=below:Star Wars\vphantom{y},below right=1.7cm
    and .5cm of sf] (sw) {};
    \node [vertex,label=below:Alec Guinness,right=2cm of sw] (alec) {};
    \node [vertex,label=above:Obi-Wan Kenobi,above=1.5cm of alec] (obi) {};

    \tikzstyle{EdgeStyle}=[->,>=stealth,thick]
    \tikzstyle{LabelStyle}=[fill=white,font=\rmfamily\footnotesize]
    \Edge[label=starredIn](nimoy)(st)
    \Edge[label=played](nimoy)(spock)
    \Edge[label=characterIn](spock)(st)
    \Edge[label=genre](st)(sf)

    \Edge[label=starredIn](alec)(sw)
    \Edge[label=played](alec)(obi)
    \Edge[label=characterIn](obi)(sw)
    \Edge[label=genre](sw)(sf)
  \end{tikzpicture}
  \caption{Sample knowledge graph. Nodes represent entities, edge labels
    represent types of relations, edges represent existing relationships.
    \label{fig:spock-example}}
\end{figure}
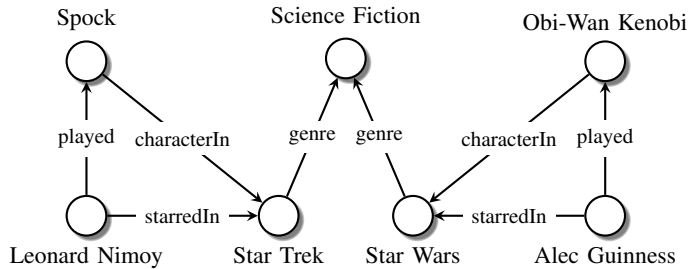

We can combine all the SPO triples together to form a multi-graph, where nodes
represent entities (all subjects and objects), and directed edges represent
relationships. The direction of an edge indicates whether entities occur as
subjects or objects, i.e., an edge points from the subject to the object.
Different relations are represented via different types of edges (also
called edge labels). This construction is called a \emph{knowledge graph} (KG),
or sometimes a \emph{heterogeneous information network}
\citep{sun_mining_2012}.) See \cref{fig:spock-example} for an example.

In addition to being a collection of facts, knowledge graphs often provide type
hierarchies (Leonard Nimoy is an actor, which is a person, which is a living
thing) and type constraints (e.g., a person can only marry another person, not a
thing).

\subsection{Open vs.\ closed world assumption}

While existing triples always encode known true relationships (facts), there are
different paradigms for the interpretation of non-existing triples:
\begin{itemize}
\item Under the \emph{closed world assumption} (CWA), non-existing triples
  indicate false relationships. For example, the fact that in
  \cref{fig:spock-example} there is no \textit{starredIn} edge from Leonard
  Nimoy to Star Wars is interpreted to mean that Nimoy definitely did not star
  in this movie.

\item Under the \emph{open world assumption} (OWA), a non-existing triple is
  interpreted as unknown, i.e., the corresponding relationship can be either
  true or false. Continuing with the above example, the missing edge is not
  interpreted to mean that Nimoy did not star in Star Wars. This more cautious
  approach is justified, since KGs are known to be very incomplete. For example,
  sometimes just the main actors in a movie are listed, not the complete cast.
  As another example, note that even the place of birth attribute, which you
  might think would be typically known, is missing for $71\%$ of all people
  included in Freebase \citep{west_knowledge_2014}.
\end{itemize}
RDF and the Semantic Web make the open-world assumption. In \cref{sec:LCWA} we
also discuss the \emph{local closed world assumption (LCWA)}, which is often
used for training relational models.

\subsection{Knowledge base construction}
Completeness, accuracy, and data quality are important parameters that determine
the usefulness of knowledge bases and are influenced by the way knowledge bases
are constructed. We can classify KB construction methods into four main groups:

\begin{itemize}
\item In \emph{curated} approaches, triples are created manually by a closed
  group of experts.

\item In \emph{collaborative} approaches, triples are created manually by an
  open group of volunteers.

\item In \emph{automated semi-structured} approaches, triples are extracted
  automatically from semi-structured text (e.g., infoboxes in Wikipedia) via
  hand-crafted rules, learned rules, or regular expressions.

\item In \emph{automated unstructured} approaches, triples are extracted
  automatically from unstructured text via machine learning and natural language
  processing techniques (see, e.g.,
  \citep{weikum_information_2010} for a review).
\end{itemize}

\begin{table}
\caption{Knowledge Base Construction Projects}\label{tab:akbc-types}
\centering
\begin{tabular}{lll}
\toprule
\textbf{Method} & \textbf{Schema} & \textbf{Examples} \\ \midrule
Curated & Yes &
          Cyc/OpenCyc~\citep{lenat_cyc:_1995},
          WordNet~\citep{miller_wordnet:_1995},\\
          & &UMLS~\citep{bodenreider_unified_2004} \\
\midrule
Collaborative & Yes &
                Wikidata~\citep{vrandecic_wikidata:_2014},
                Freebase~\citep{bollacker_freebase:_2008} \\
\midrule
Auto. Semi-Structured & Yes &
                        YAGO~\citep{suchanek_yago:_2007,hoffart_yago2:_2013},
                        DBPedia~\citep{auer_dbpedia:_2007},\\
                        && Freebase~\citep{bollacker_freebase:_2008} \\
\midrule
Auto. Unstructured & Yes &
                     Knowledge Vault~\citep{dong_knowledge_2014},
                     NELL~\citep{carlson_toward_2010},\\
                     && PATTY~\citep{nakashole_patty:_2012},
                     PROSPERA~\citep{nakashole_scalable_2011}, \\
                     && DeepDive/Elementary~\citep{niu_elementary:_2012}\\
\midrule
Auto. Unstructured & No &
ReVerb~\citep{fader_identifying_2011}, OLLIE~\citep{schmitz_open_2012},\\
&&  PRISMATIC~\citep{fan_prismatic:_2010} \\
\bottomrule
\end{tabular}
\end{table}

Construction of curated knowledge bases typically leads to highly accurate
results, but this technique does not scale well due to its dependence on human
experts. Collaborative knowledge base construction, which was used to build
Wikipedia and Freebase, scales better but still has some limitations. For
instance, as mentioned previously, the place of birth attribute is missing for
$71\%$ of all people included in Freebase, even though this is a mandatory
property of the schema \citep{west_knowledge_2014}. Also, a recent study
\citep{suh_singularity_2009} found that the growth of Wikipedia has been slowing
down. Consequently, automatic knowledge base construction methods have been
gaining more attention.

Such methods can be grouped into two main approaches. The first approach
exploits semi-structured data, such as Wikipedia infoboxes, which has led to
large, highly accurate knowledge graphs such as
YAGO~\citep{suchanek_yago:_2007,hoffart_yago2:_2013} and
DBpedia~\citep{auer_dbpedia:_2007}. The accuracy (trustworthiness) of facts in
such automatically created KGs is often still very high. For instance, the
accuracy of YAGO2 has been estimated\footnote{For detailed statistics see \url{http://www.mpi-inf.mpg.de/departments/databases-and-information-systems/research/yago-naga/yago/statistics/}}
to be over 95\% through manual inspection of sample
facts~\citep{biega_inside_2013}, and the accuracy of
Freebase~\citep{bollacker_freebase:_2008} was estimated to be
99\%\footnote{\url{http://thenoisychannel.com/2011/11/15/cikm-2011-industry-event-john-giannandrea-on-freebase-a-rosetta-stone-for-entities}}.
However, semi-structured text still covers only a small fraction of the
information stored on the Web, and completeness (or coverage) is another
important aspect of KGs. Hence the second approach tries to ``read the Web'',
extracting facts from the natural language text of Web pages. Example projects
in this category include NELL \citep{carlson_toward_2010} and the Knowledge
Vault \citep{dong_knowledge_2014}. In Section~\ref{sec:kv}, we show how we can
reduce the level of ``noise'' in such automatically extracted facts by using the
knowledge from existing, high-quality repositories.

KGs, and more generally KBs, can also be classified based on whether they employ
a fixed or open lexicon of entities and relations. In particular, we distinguish
two main types of KBs:
\begin{itemize}
\item In \emph{schema-based} approaches, entities and relations are represented
  via globally unique identifiers and all possible relations are predefined in a
  fixed vocabulary. For example, Freebase might represent the fact that Barack
  Obama was born in Hawaii using the triple (/m/02mjmr, /people/person/born-in,
  /m/03gh4), where /m/02mjmr is the unique machine ID for Barack Obama.

\item In \emph{schema-free} approaches, entities and relations are identified
  using open information extraction (OpenIE) techniques
  \citep{etzioni_open_2011}, and represented via normalized but not
  disambiguated strings (also referred to as surface names). For example, an
  OpenIE system may contain triples such as (``Obama'', ``born in'',
  ``Hawaii''), (``Barack Obama'', ``place of birth'', ``Honolulu''), etc. Note
  that it is not clear from this representation whether the first triple refers
  to the same person as the second triple, nor whether ``born in'' means the
  same thing as ``place of birth''. This is the main disadvantage of OpenIE
  systems.
\end{itemize}

\begin{table}
  \caption{Size of some schema-based knowledge bases}\label{tab:kg-sizes}
  \centering
  \begin{tabular}{lrrr}
    \toprule
    & \multicolumn{3}{c}{Number of} \\
     \cmidrule(l){2-4}
    \textbf{Knowledge Graph} & \textbf{Entities} &
    \textbf{Relation Types} & \textbf{Facts} \\
    \cmidrule(r){1-1}\cmidrule(lr){2-2}\cmidrule(lr){3-3}\cmidrule(l){4-4}
    Freebase\tablefootnote{Non-redundant triples, see~\citep[Table 1]{dong_knowledge_2014}} & 40 M & 35,000 & 637 M \\
    Wikidata\tablefootnote{Last published numbers:
    \url{https://tools.wmflabs.org/wikidata-todo/stats.php} and
    \url{https://www.wikidata.org/wiki/Category:All_Properties}} & 18 M & 1,632 & 66 M \\
    DBpedia (en)\tablefootnote{English content, Version 2014 from \url{http://wiki.dbpedia.org/data-set-2014}}  & 4.6 M & 1,367 & 538 M \\
    YAGO2 \tablefootnote{See~\citep[Table 5]{hoffart_yago2:_2013}}& 9.8 M & 114 & 447 M \\
    Google Knowledge Graph\tablefootnote{Last published numbers: \url{http://insidesearch.blogspot.de/2012/12/get-smarter-answers-from-knowledge_4.html}} & 570 M & 35,000 & 18,000 M \\
    \bottomrule
  \end{tabular}
\end{table}

\Cref{tab:akbc-types} lists current knowledge base construction projects
classified by their creation method and data schema.
In this paper, we will only focus on schema-based KBs.
\Cref{tab:kg-sizes} shows a selection of such KBs and their sizes.

\subsection{Uses of knowledge graphs}
Knowledge graphs provide semantically structured information that is
interpretable by computers --- a property that is regarded as an important
ingredient to build more intelligent machines~\citep{lenat_thresholds_1991}.
Consequently, knowledge graphs are already powering multiple ``Big Data''
applications in a variety of commercial and scientific domains. A prime example
is the integration of Google's Knowledge Graph, which currently stores 18
billion facts about 570 million entities, into the results of Google's search
engine \cite{singhal_introducing_2012}. The Google Knowledge Graph is used to
identify and disambiguate entities in text, to enrich search results with
semantically structured summaries, and to provide links to related entities in
exploratory search. (Microsoft has a similar KB, called Satori, integrated with
its Bing search engine~\citep{qian_understand_2013}.)

Enhancing search results with semantic information from knowledge graphs can be
seen as an important step to transform text-based search engines into
semantically aware question answering services. Another prominent example
demonstrating the value of knowledge graphs is IBM's question answering system
Watson, which was able to beat human experts in the game of \emph{Jeopardy!}.
Among others, this system used YAGO, DBpedia, and Freebase as its sources of
information~\citep{ferrucci_building_2010}. Repositories of structured knowledge
are also an indispensable component of digital assistants such as Siri, Cortana,
or Google Now.

Knowledge graphs are also used in
several specialized domains.  For instance,
Bio2RDF~\cite{belleau_bio2rdf:_2008},
Neurocommons~\cite{ruttenberg_life_2009}, and
LinkedLifeData~\cite{momtchev_expanding_2009} are knowledge graphs
that integrate multiple sources of biomedical information.
These have been used for
question answering and decision support in the life sciences.

\subsection{Main tasks in knowledge graph construction and curation}
In this section, we review a number of typical KG tasks. \medskip

\emph{Link prediction} is concerned with predicting the existence (or
probability of correctness) of (typed) edges in the graph (i.e., triples). This
is important since existing knowledge graphs are often missing many facts, and
some of the edges they contain are incorrect~\citep{angeli_philosophers_2013}.
In the context of knowledge graphs, link prediction is also referred to as
\emph{knowledge graph completion}. For example, in
Figure~\ref{fig:spock-example}, suppose the \textit{characterIn} edge from
\textit{Obi-Wan Kenobi} to \textit{Star Wars} were missing; we might be able to
predict this missing edge, based on the structural similarity between this part
of the graph and the part involving \textit{Spock} and \textit{Star Trek}. It
has been shown that relational models that take the relationships of entities
into account can significantly outperform non-relational machine learning
methods for this task (e.g., see \citep{taskar_link_2004,getoor_link_2005}).

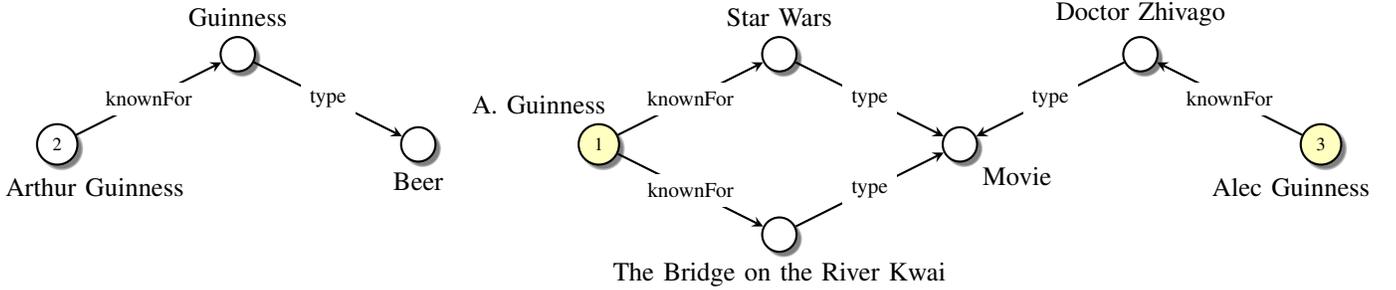
\begin{figure*}[tb]
  \centering
  \begin{tikzpicture}[scale=1.2,baseline,thick]
    \GraphInit[vstyle=Classic]
    \tikzset{vertex/.style =
      {draw=black,shape=circle,fill=white,minimum size=13pt,circular
        drop shadow}}
    \node at (4,0)[vertex,label=below:The Bridge on the River Kwai] (pub1) {};
    \node at (4,2)[vertex,label=above:Star Wars] (pub2) {};
    \node at (8,2)[vertex,label={Doctor Zhivago}] (pub3) {};
    \node at (-2,2)[vertex,label=above:Guinness] (pub4)
    {};

    \node at (6,1)[vertex,label=below right:Movie] (ven1) {};
    \node at (0,1)[vertex,label=below:Beer] (ven2) {};

    \node at (2,1)[vertex,fill=ccls32,label={[xshift=-0.8cm]A. Guinness}] (aut1) {\scriptsize 1};
    \node at
    (-4,1)[vertex,label={[yshift=-1.1cm,xshift=0.5cm]Arthur Guinness}]
    (aut2) {\scriptsize 2};
    \node at
    (10,1)[vertex,fill=ccls32,label={[yshift=-1.1cm,xshift=-0.4cm]Alec Guinness}]
    (aut3) {\scriptsize 3};

    \tikzstyle{EdgeStyle}=[->,>=stealth,thick]
    \tikzstyle{LabelStyle}=[fill=white,pos=0.5]
    \Edge[label={\footnotesize{knownFor}}](aut1)(pub1)
    \Edge[label={\footnotesize{knownFor}}](aut1)(pub2)
    \Edge[label={\footnotesize{knownFor}}](aut3)(pub3)
    \Edge[label={\footnotesize{knownFor}}](aut2)(pub4)
    \Edge[label={\footnotesize{type}}](pub1)(ven1)
    \Edge[label={\footnotesize{type}}](pub2)(ven1)
    \Edge[label={\footnotesize{type}}](pub3)(ven1)
    \Edge[label={\footnotesize{type}}](pub4)(ven2)
  \end{tikzpicture}
  \caption{Example of entity resolution in a toy knowledge graph. In
    this example, nodes 1 and 3 refer to the identical entity, the actor Alec Guinness.
    Node 2 on the other hand refers to Arthur Guinness, the founder of the Guinness brewery.
    The surface name of node 2 (``A. Guinness'') alone would not
    be sufficient to perform a correct matching as it could refer to both Alec
    Guinness and Arthur Guinness. However, since links in the
    graph reveal the occupations of the persons, a relational approach can perform the correct matching.\label{fig:er-example}}
\end{figure*}

\emph{Entity resolution} (also known as record
linkage~\citep{newcombe_automatic_1959}, object
identification~\citep{tejada_learning_2001}, instance
matching~\citep{rahm_survey_2001}, and
de-duplication~\citep{culotta_joint_2005}) is the problem of identifying which
objects in relational data refer to the same underlying entities. See
\cref{fig:er-example} for a small example. In a relational setting, the
decisions about which objects are assumed to be identical can propagate through
the graph, so that matching decisions are made \emph{collectively} for all
objects in a domain rather than independently for each object pair (see, for
example,
\citep{singla_entity_2006,bhattacharya_collective_2007,whang_joint_2012}). In
schema-based automated knowledge base construction, entity resolution can be
used to match the extracted surface names to entities stored in the knowledge
graph.

\emph{Link-based clustering} extends feature-based clustering to a relational
learning setting and groups entities in relational data based on their
similarity. However, in link-based clustering, entities are not only grouped by
the similarity of their features but also by the similarity of their links. As
in entity resolution, the similarity of entities can propagate through the
knowledge graph, such that relational modeling can add important information for
this task. In social network analysis, link-based clustering is also known as
community detection~\citep{fortunato_community_2010}.


\section{Statistical Relational Learning for Knowledge Graphs}
\label{sec:srl}

\emph{Statistical Relational Learning} is concerned with the creation
of statistical models for relational data.  In the following sections
we discuss how statistical relational learning can be applied to
knowledge graphs. We will assume that all the entities and (types of)
relations in a knowledge graph are known.  (We discuss extensions of
this assumption in \cref{sec:newEntities}).  However, triples are
assumed to be incomplete and noisy; entities and relation types may
contain duplicates.

\subsubsection*{Notation}
\label{sec:notation}

Before proceeding, let us define our mathematical notation. (Variable names will
be introduced later in the appropriate sections.) We denote scalars by lower
case letters, such as $a$; column vectors (of size $N$) by bold lower case
letters, such as $\va$; matrices (of size $N_1 \times N_2$) by bold upper case
letters, such as $\mA$; and tensors (of size $N_1 \times N_2 \times N_3$) by
bold upper case letters with an underscore, such as $\tA$. We denote the $k$'th
``frontal slice'' of a tensor $\tA$ by $\mA_k$ (which is a matrix of size $N_1
\times N_2$), and the $(i,j,k)$'th element by $a_{ijk}$ (which is a scalar). We
use $[\va; \vb]$ to denote the vertical stacking of vectors $\va$ and $\vb$,
i.e., $[\va; \vb] = \begin{pmatrix} \va \\ \vb \end{pmatrix}$. We can convert a
matrix $\mA$ of size $N_1 \times N_2$ into a vector $\va$ of size $N_1N_2$ by
stacking all columns of $\mA$, denoted $\va=\vect{\mA}$. The inner (scalar)
product of two vectors (both of size $N$) is defined by $\va^\transp \vb =
\sum_{i=1}^N a_i b_i$. The tensor (Kronecker) product of two vectors (of size
$N_1$ and $N_2$) is a vector of size $N_1N_2$ with entries $\va \kron \vb
= \begin{pmatrix}a_1 \vb\\ \vdots \\a_{N_1} \vb\end{pmatrix}$. Matrix
multiplication is denoted by $\mA \mB$ as usual. We denote the $L_2$ norm of a
vector by $||\va||_2 = \sqrt{\sum_i a_i^2}$, and the Frobenius norm of a matrix
by $||\mA||_F=\sqrt{\sum_i \sum_j a_{ij}^2}$. We denote the vector of all ones
by $\vone$, and the identity matrix by $\mI$.

\subsection{Probabilistic knowledge graphs}
We now introduce some mathematical background so we can more formally
define statistical models for knowledge graphs.

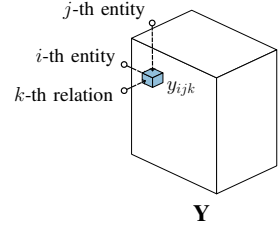
\begin{figure}[t]
  \centering
  \hspace{-3em}\resizebox{.4\linewidth}{!}{\begin{tikzpicture}[y=0.80pt, x=0.8pt,yscale=-1, inner sep=0pt, outer sep=0pt,scale=0.7]
\begin{scope}[shift={(329.23291,-382.9165)}]
  \begin{scope}[shift={(-553.24566,-46.35037)}]
    \begin{scope}[shift={(90.49286,-355.07313)},draw=black,fill=white,line join=round,line cap=butt,line width=0.800pt]
      \path[draw,fill] (227.9600,836.2200) -- (227.9600,1048.8200) --
        (383.7300,1121.4580) -- (383.7300,908.8580) -- (227.9600,836.2200) -- cycle;
      \path[draw,fill] (227.9600,836.2200) -- (338.1400,784.8400) --
        (493.9100,857.4800) -- (383.7300,908.8600) -- cycle;
      \path[shift={(0,308.2677)},draw,fill] (383.7300,813.1900) -- (493.9100,761.8120)
        -- (493.9100,549.2120) -- (383.7300,600.5900) -- cycle;
    \end{scope}
  \end{scope}
  \path[draw=black,dash pattern=on 4.80pt off 1.20pt,line join=miter,line
    cap=butt,miter limit=4.00,line width=1.200pt,o->] (-254.0433,480.3123) -- (-208.1249,501.5300);
  \begin{scope}[shift={(-331.70385,40.21882)},draw=black,fill=ccls5,line join=round]
    \path[shift={(-235.40819,370.25803)},draw=black,fill=ccls5,line
      join=round,line cap=butt,line width=0.800pt] (351.7989,87.9086) --
      (351.7989,105.6252) -- (370.7955,114.4834) -- (370.7955,96.7669) -- cycle;
    \path[shift={(-235.40819,370.25803)},draw=black,fill=ccls5,line
      join=round,line cap=butt,line width=0.800pt] (351.7989,87.9086) --
      (370.7955,79.0504) -- (389.7921,87.9086) -- (370.7955,96.7669) -- cycle;
    \path[shift={(-235.40819,370.25803)},draw=black,fill=ccls5,line
      join=round,line cap=butt,line width=0.800pt] (370.7955,96.7669) --
      (370.7955,114.4834) -- (389.7921,105.6252) -- (389.7921,87.9086) -- cycle;
  \end{scope}
  \path[fill=black] (-410.03314,485.31232) node[above right,align=right] (text4469) {\huge $i$-th entity};
  \path[draw=black,dash pattern=on 4.80pt off 1.20pt,line join=miter,line
    cap=butt,miter limit=4.00,line width=1.200pt,o->] (-197.0535,400.5879) -- (-196.3392,495.8157);
  \path[draw=black,dash pattern=on 4.80pt off 1.20pt,line join=miter,line
    cap=butt,miter limit=4.00,line width=1.200pt,o->] (-254.0433,533.4619) -- (-208.4820,512.2443);
  \path[fill=black] (-360.05345,399.72964) node[above right] (text5781) {\huge $j$-th
    entity};
  \path[fill=black] (-450.02979,550.32019) node[above right,align=right] (text5787) {\huge $k$-th relation};
  \path[fill=black] (-122.16412,769.6377) node[above right] (text3077) {\Huge $\ten{Y}$};
  \path[fill=black] (-170.16412,540.6377) node[above right] (text3078) {\huge $y_{ijk}$};
\end{scope}
\end{tikzpicture}}
  \caption{Tensor representation of binary relational
    data\label{fig:adjtensor}.}
\end{figure}

Let $\Set{E} = \{e_1,\dots ,e_{\Nentities}\}$ be the set of all entities and
${\Set{R} = \{r_1, \dots, r_{\Nreln}\}}$ be the set of all relation types in a
knowledge graph. We model each \emph{possible} triple $x_{ijk} = (e_i, r_k,
e_j)$ over this set of entities and relations as a binary random variable
$y_{ijk} \in \{0,1\}$ that indicates its existence. All possible triples in
$\Set{E} \times \Set{R} \times \Set{E}$ can be grouped naturally in a
third-order tensor (three-way array) ${\tY \in {\{0,1\}}^{\Nentities \times
    \Nentities \times \Nreln}}$, whose entries are set such that
\[
y_{ijk} = \begin{cases}
  1,& \text{if the triple } (e_i, r_k, e_j) \text{ exists} \\
  0, & \text{otherwise.}
\end{cases}
\]
We will refer to this construction as an
\emph{adjacency tensor} (cf.\ \Cref{fig:adjtensor}).
Each possible realization of $\tY$ can be interpreted as a possible world. To
derive a model for the entire knowledge graph, we are then interested in
estimating the joint distribution $\prob(\tY)$, from a subset $\Set{D} \subseteq
\Set{E} \times \Set{R} \times \Set{E} \times \{0,1\}$ of observed triples. In
doing so, we are estimating a probability distribution over possible worlds,
which allows us to predict the probability of triples based on the state of the
entire knowledge graph. While $y_{ijk} = 1$ in adjacency tensors indicates the
existence of a triple, the interpretation of $y_{ijk} = 0$ depends on whether
the open world, closed world, or local-closed world assumption is made. For
details, see~\cref{sec:LCWA}.

Note that the size of $\tY$ can be enormous for large knowledge graphs. For
instance, in the case of Freebase, which currently consists of over $40$ million
entities and $35,000$ relations, the number of possible triples $|\Set{E} \times
\Set{R} \times \Set{E}|$ exceeds $10^{19}$ elements. Of course, type constraints
reduce this number considerably.

Even amongst the syntactically valid triples, only a tiny fraction are likely to
be true. For example, there are over 450,000 thousands actors and over 250,000
movies stored in Freebase. But each actor stars only in a small number of
movies. Therefore, an important issue for SRL on knowledge graphs is how to deal
with the large number of possible relationships while efficiently exploiting the
sparsity of relationships. Ideally, a relational model for large-scale knowledge
graphs should scale at most linearly with the data size, i.e., linearly in the
number of entities $\Nentities$, linearly in the number of relations $\Nreln$,
and linearly in the number of \emph{observed} triples $|\Set{D}|=\Nsamples$.

\subsection{Statistical properties of knowledge graphs}
\label{sec:kg-properties}

Knowledge graphs typically adhere to some deterministic rules, such as type
constraints and transitivity (e.g., if Leonard Nimoy was born in Boston, and
Boston is located in the USA, then we can infer that Leonard Nimoy was born in
the USA). However, KGs have typically also various ``softer'' statistical
patterns or regularities, which are not universally true but nevertheless have
useful predictive power.

One example of such statistical pattern is known as \emph{homophily}, that is,
the tendency of entities to be related to other entities with similar
characteristics. This has been widely observed in various social
networks~\citep{newman_structure_2001,liben-nowell_link-prediction_2007}. For
example, US-born actors are more likely to star in US-made movies. For
multi-relational data (graphs with more than one kind of link), homophily has
also been referred to as \emph{autocorrelation}~\citep{jensen_linkage_2002}.

Another statistical pattern is known as \emph{block structure}. This refers to
the property where entities can be divided into distinct groups (blocks), such
that all the members of a group have similar relationships to members of other
groups~\citep{holland_stochastic_1983,anderson_building_1992,hoff_modeling_2008}.
For example, we can group some actors, such as Leonard Nimoy and Alec Guinness,
into a science fiction actor block, and some movies, such as Star Trek and Star
Wars, into a science fiction movie block, since there is a high density of links
from the scifi actor block to the scifi movie block.

Graphs can also exhibit \emph{global and long-range statistical dependencies},
i.e., dependencies that can span over chains of triples and involve different
types of relations. For example, the citizenship of Leonard Nimoy (USA) depends
statistically on the city where he was born (Boston), and this dependency
involves a path over multiple entities (Leonard Nimoy, Boston, USA) and
relations (\textit{bornIn, locatedIn, citizenOf}). A distinctive feature of
relational learning is that it is able to exploit such patterns to create richer
and more accurate models of relational domains.

When applying statistical models to incomplete knowledge graphs, it should be
noted that the distribution of facts in such KGs can be skewed. For instance,
KGs that are derived from Wikipedia will inherit the skew that exists in
distribution of facts in Wikipedia itself.\footnote{As an example, there are
  currently 10,306 male and 7,586 female American actors listed in Wikipedia,
  while there are only 1,268 male and 1,354 female Indian, and 77 male and no
  female Nigerian actors. India and Nigeria, however, are the largest and second
  largest film industries in the world.} Statistical models as discussed in the
following sections can be affected by such biases in the input data and need to
be interpreted accordingly.

\subsection{Types of SRL models}
\label{sec:SRLtypes}

As we discussed, the presence or absence of certain triples in relational data
is correlated with (i.e., predictive of) the presence or absence of certain
other triples. In other words, the random variables $y_{ijk}$ are correlated
with each other. We will discuss three main ways to model these correlations:
\begin{description}
\item[M1)] Assume all $y_{ijk}$ are conditionally independent given latent
  features associated with subject, object and relation type and additional
  parameters (\emph{latent feature models})
\item[M2)] Assume all $y_{ijk}$ are conditionally independent given observed
  graph features and additional parameters (\emph{graph feature models})
\item[M3)] Assume all $y_{ijk}$ have local interactions (\emph{Markov Random
    Fields})
\end{description}
In what follows we will mainly focus on M1 and M2 and their combination; M3 will
be the topic of Section~\ref{sec:mln}.

The model classes M1 and M2 predict the existence of a triple $x_{ijk}$ via a
score function $f(x_{ijk};\Theta)$ which represents the model's confidence that
a triple exists given the parameters $\Theta$. The conditional independence
assumptions of M1 and M2 allow the probability model to be written as follows:
\begin{equation}
	\prob(\tY | \calD, \Theta ) =
	\prod_{i=1}^{\Nentities}
  \prod_{j=1}^{\Nentities}
  \prod_{k=1}^{\Nreln}
  \operatorname{Ber}(y_{ijk}\,|\,\sigma(f(x_{ijk}; \Theta)))
	\label{eq:lfm:joint}
\end{equation}
where $\sigma(u)=1/(1+e^{-u})$ is the sigmoid (logistic) function, and
\begin{equation}
  \operatorname{Ber}(y | p) =
\left\{ \begin{array}{ll}
 p &\mbox{if $y=1$} \\
 1-p &\mbox{if $y=0$}
\end{array} \right.
\end{equation}
is the Bernoulli distribution.

We will refer to models of the form \cref{eq:lfm:joint} as \emph{probabilistic
  models}. In addition to probabilistic models, we will also discuss models
which optimize $f(\cdot)$ under other criteria, for instance models which
maximize the margin between existing and non-existing triples. We will refer to
such models as \emph{score-based models}. If desired, we can derive
probabilities for score-based models via Platt
scaling~\citep{platt_probabilities_1999}.

There are many different methods for defining $f(\cdot)$. In the following
\cref{sec:lfm,sec:observable,sec:combining,sec:MRF} we will discuss different
options for all model classes. In \cref{sec:training} we will furthermore
discuss aspects of how to train these models on knowledge graphs.


\section{Latent Feature Models}
\label{sec:lfm}

In this section, we assume that the variables $y_{ijk}$ are conditionally
independent given a set of global latent features and parameters, as in
Equation~\ref{eq:lfm:joint}. We discuss various possible forms for the score
function $f(x;\Theta)$ below. What all models have in common is that they
explain triples via latent features of entities (This is justified via various
theoretical arguments \cite{Orbanz2015}). For instance, a possible explanation
for the fact that Alec Guinness received the Academy Award is that he is a good
actor. This explanation uses latent features of entities (being a good actor) to
explain observable facts (Guinness receiving the Academy Award). We call these
features ``latent'' because they are not directly observed in the data. One task
of all latent feature models is therefore to infer these features automatically
from the data.

In the following, we will denote the latent feature representation of an entity
$e_i$ by the vector $\ve_i \in \R^{\dimentities}$ where $\dimentities$ denotes
the number of latent features in the model. For instance, we could model that
Alec Guinness is a good actor and that the Academy Award is a prestigious award
via the vectors
\[
{\vec{e}}_{\text{Guinness}} = \begin{bmatrix}
  0.9 \\ {0.2}
\end{bmatrix},\quad
{\vec{e}}_{\text{AcademyAward}} = \begin{bmatrix}
  0.2 \\ {0.8}
\end{bmatrix}
\]
where the component $e_{i1}$ corresponds to the latent feature \emph{Good Actor}
and $e_{i2}$ correspond to \emph{Prestigious Award}. (Note that, unlike this
example, the latent features that are inferred by the following models are
typically hard to interpret.)

The key intuition behind relational latent feature models is that the
relationships between entities can be derived from interactions of their latent
features. However, there are many possible ways to model these interactions, and
many ways to derive the existence of a relationship from them. We discuss
several possibilities below. See Table~\ref{tab:notation} for a summary of the
notation.

\begin{table}
\caption{Summary of the notation.}
\label{tab:notation}
\centering
\hspace{-1em}
\begin{tabular}{lll}
  \toprule
  \multicolumn{3}{c}{\textbf{Relational data}}\\
  Symbol & Meaning \\
  \midrule
  $\Nentities$ & Number of entities \\
  $\Nreln$ & Number of relations \\
  $\Nsamples$ & Number of training examples \\
  $e_i$ & \multicolumn{2}{l}{$i$-th entity in the dataset (e.g., \textit{LeonardNimoy})} \\
  $r_k$ & \multicolumn{2}{l}{$k$-th relation in the dataset (e.g., \textit{bornIn})}\\
  $\calD^+$ & Set of observed positive triples \\
  $\calD^-$ & Set of observed negative triples \\
  \midrule
  \multicolumn{3}{c}{\textbf{Probabilistic Knowledge Graphs}}\\
  Symbol & Meaning & Size\\
  \midrule
  $\tY$ & (Partially observed) labels for all triples & $\Nentities \times \Nentities \times \Nreln$\\
  $\tF$ & Score for all possible triples & $\Nentities \times \Nentities \times \Nreln$ \\
  $\mY_k$ & Slice of $\tY$ for relation $r_k$ & $\Nentities \times \Nentities$\\
  $\mF_k$ & Slice of $\tF$ for relation $r_k$ & $\Nentities \times
                                                \Nentities$\\
  \midrule
  \multicolumn{3}{c}{\textbf{Graph and Latent Feature Models}}\\
  Symbol & Meaning \\
  \midrule
  $\vphi_{ijk}$ & \multicolumn{2}{l}{Feature vector representation of triple $(e_i, r_k,e_j)$} \\
  $\vw_k$ & \multicolumn{2}{l}{Weight vector to derive scores for relation $k$} \\
  $\Theta$ & Set of all parameters of the model \\
  $\sigma(\cdot)$ & Sigmoid (logistic) function  \\
  \midrule
  \multicolumn{3}{c}{\textbf{Latent Feature Models}}\\
  Symbol & Meaning & Size\\
  \midrule
  $\dimentities$ & Number of latent features for entities \\
  $\dimreln$ & Number of latent features for relations \\
  $\ve_i$ & Latent feature repr. of entity $e_i$ & $\dimentities$ \\
  $\vr_k$ & Latent feature repr. of relation $r_k$ & $\dimreln$ \\
  $H_a$ & Size of $\vh_a$ layer \\
  $H_b$ & Size of $\vh_b$ layer \\
  $H_c$ & Size of $\vh_c$ layer \\
$\mE$ & Entity embedding matrix & $\Nentities \times \dimentities$ \\
$\mW_k$ & Bilinear weight matrix for relation $k$ & $H_e \times H_e$ \\
$\mA_k$ & Linear feature map for pairs of entities & $(2
                                                     \dimentities) \times H_a$ \\
& for relation $r_k$ \\
$\mC$ & Linear feature map for triples & $ (2 \dimentities +
                                         \dimreln) \times H_c  $ \\
\bottomrule
\end{tabular}
\end{table}

\subsection{RESCAL: A bilinear model}
\label{sec:rescal}

RESCAL~\cite{nickel_three-way_2011,nickel_factorizing_2012,nickel_tensor_2013}
is a relational latent feature model which explains triples via pairwise
interactions of latent features. In particular, we model the score of a triple
$x_{ijk}$ as
\begin{equation}
  \label{eq:rescal}
  f_{ijk}^{\text{RESCAL}} \mdef \ve_i^\transp \mat{W}_k \ve_j
 = \sum_{a=1}^{H_e} \sum_{b=1}^{H_e} w_{abk} e_{ia} e_{jb}
\end{equation}
where $\mat{W}_k \in \R^{\dimentities \times \dimentities}$ is a weight matrix
whose entries $w_{abk}$ specify how much the latent features $a$ and $b$
interact in the $k$-th relation. We call this a bilinear model, since it
captures the interactions between the two entity vectors using multiplicative
terms. For instance, we could model the pattern that \emph{good actors are
  likely to receive prestigious awards} via a weight matrix such as
\[
\mat{W}_{\text{receivedAward}} = \begin{bmatrix}
  0.1 & \colorbox{ccls33!50}{0.9} \\
  0.1 & 0.1
\end{bmatrix} .
\]
In general, we can model block structure patterns via the magnitude of entries
in $\mat{W}_k$, while we can model homophily patterns via the magnitude of its
diagonal entries. Anti-correlations in these patterns can be modeled via
negative entries in $\mat{W}_k$.

Hence, in \Cref{eq:rescal} we compute the score of a triple $x_{ijk}$ via the
weighted sum of all pairwise interactions between the latent features of the
entities $e_i$ and $e_j$. The parameters of the model are $\Theta =
\{\{\ve_i\}_{i=1}^{\Nentities}, \{\mat{W}_k\}_{k=1}^{\Nreln}\}$. During training
we \emph{jointly} learn the latent representations of entities and how the
latent features interact for particular relation types.

In the following, we will discuss further important properties of the model for
learning from knowledge graphs.

\subsubsection*{Relational learning via shared representations}
In equation~\eqref{eq:rescal}, entities have the same latent representation
regardless of whether they occur as subjects or objects in a relationship.
Furthermore, they have the same representation over all different relation
types. For instance, the $i$-th entity occurs in the triple $x_{ijk}$ as the
subject of a relationship of type $k$, while it occurs in the triple $x_{piq}$
as the object of a relationship of type $q$. However, the predictions $f_{ijk} =
\ve_i^\transp \mat{W}_k \ve_j$ and $f_{piq} = \ve_p^\transp \mat{W}_q \ve_i$
both use the same latent representation $\ve_i$ of the $i$-th entity. Since all
parameters are learned jointly, these shared representations permit to propagate
information between triples via the latent representations of entities and the
weights of relations. This allows the model to capture global dependencies in
the data.

\subsubsection*{Semantic embeddings}
The shared entity representations in RESCAL capture also the similarity of
entities in the relational domain, i.e., that \emph{entities are similar if they
  are connected to similar entities via similar
  relations}~\citep{nickel_tensor_2013}. For instance, if the representations of
$\ve_i$ and $\ve_p$ are similar, the predictions $f_{ijk}$ and $f_{pjk}$ will
have similar values. In return, entities with many similar observed
relationships will have similar latent representations. This property can be
exploited for entity resolution and has also enabled large-scale hierarchical
clustering on relational
data~\cite{nickel_three-way_2011,nickel_factorizing_2012}. Moreover, since
relational similarity is expressed via the similarity of vectors, the latent
representations $\ve_i$ can act as proxies to give non-relational machine
learning algorithms such as $k$-means or kernel methods access to the relational
similarity of entities.

\subsubsection*{Connection to tensor factorization}
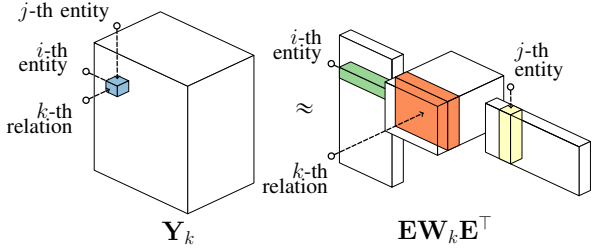
\begin{figure}[t]
  \vspace{1.2em}
  \centering
  \resizebox{0.9\columnwidth}{!}{\resizebox{\linewidth}{!}{
\begin{tikzpicture}[y=0.80pt, x=0.8pt,yscale=-1, inner sep=0pt, outer sep=0pt,scale=0.7]
\begin{scope}[shift={(329.23291,-382.9165)}]
  \begin{scope}[shift={(-176.0418,-715.0068)},draw=black,fill=white,line join=round]
    \path[draw=black,fill=white,line join=round,line cap=butt,line width=0.800pt]
      (349.5378,1137.4016) -- (368.5344,1128.5433) -- (455.9188,1169.2913) --
      (436.9222,1178.1496) -- cycle;
    \path[draw=black,fill=white,line join=round,line cap=butt,line width=0.800pt]
      (436.9222,1178.1496) -- (455.9188,1169.2913) -- (455.9188,1381.8898) --
      (436.9222,1390.7481) -- cycle;
    \path[draw=black,fill=white,line join=round,line cap=butt,line width=0.800pt]
      (349.5378,1137.4016) -- (436.9222,1178.1496) -- (436.9222,1390.7480) --
      (349.5378,1350.0000) -- cycle;
  \end{scope}
  \path[draw=black,dash pattern=on 4.80pt off 1.20pt,line join=miter,line
    cap=butt,miter limit=4.00,line width=1.200pt,o->] (154.3839,467.0249) -- (181.3057,480.6664);
  \begin{scope}[shift={(-0.89286,35.35714)}]
    \begin{scope}[shift={(0.82967,-17.83878)},draw=black,fill=ccls33,line join=round]
      \path[draw=black,fill=ccls33,line join=round,line cap=butt,line width=0.800pt]
        (173.4699,458.4479) -- (173.4699,476.1644) -- (260.8543,516.9124) --
        (260.8543,499.1959) -- cycle;
      \path[shift={(172.86675,413.36317)},draw=black,fill=ccls33,line join=round,line
        cap=butt,line width=0.800pt] (0.5138,44.8035) -- (19.5104,35.9452) --
        (106.8949,76.6932) -- (87.8982,85.5515) -- cycle;
      \path[shift={(172.86675,413.36317)},draw=black,fill=ccls33,line join=round,line
        cap=butt,line width=0.800pt] (87.8982,85.5515) -- (87.8982,103.2680) --
        (106.8949,94.4098) -- (106.8949,76.6932) -- cycle;
    \end{scope}
    \begin{scope}[shift={(90.42463,46.3035)},draw=black,fill=white,opacity=0.000,line join=round,transparency group]
      \path[draw=black,fill=white,line join=round,line cap=butt,line width=0.800pt]
        (151.9729,396.8504) -- (151.9729,609.4488) -- (170.9695,618.3071) --
        (170.9695,405.7086) -- cycle;
      \path[draw=black,fill=white,line join=round,line cap=butt,line width=0.800pt]
        (151.9729,396.8504) -- (170.9695,387.9921) -- (189.9662,396.8504) --
        (170.9695,405.7086) -- cycle;
      \path[draw=black,fill=white,line join=round,line cap=butt,line width=0.800pt]
        (170.9695,618.3071) -- (189.9662,609.4488) -- (189.9662,396.8504) --
        (170.9695,405.7086) -- cycle;
    \end{scope}
  \end{scope}
  \begin{scope}[shift={(21.523,-335.8729)},draw=black,fill=white,line join=round]
    \path[draw=black,fill=white,line join=round,line cap=butt,line width=0.800pt]
      (227.9594,836.2205) -- (227.9594,917.7165) -- (315.3438,958.4646) --
      (315.3438,876.9685) -- cycle;
    \path[draw=black,fill=white,line join=round,line cap=butt,line width=0.800pt]
      (227.9594,836.2205) -- (338.1398,784.8425) -- (425.5242,825.5905) --
      (383.7317,908.8582) -- cycle;
    \path[shift={(0,308.2677)},draw=black,fill=white,line join=round,line
      cap=butt,line width=0.800pt] (315.3438,650.1969) -- (425.5242,598.8189) --
      (425.5242,517.3228) -- (315.3438,568.7008) -- cycle;
  \end{scope}
  \begin{scope}[shift={(-77.2594,-587.4477)},draw=black,fill=white,line join=round]
    \path[draw=black,fill=white,line join=round,line cap=butt,line width=0.800pt]
      (493.9120,1123.2283) -- (512.9086,1114.3701) -- (668.6809,1187.0079) --
      (649.6843,1195.8661) -- cycle;
    \path[draw=black,fill=white,line join=round,line cap=butt,line width=0.800pt]
      (649.6843,1195.8661) -- (668.6809,1187.0079) -- (668.6809,1268.5039) --
      (649.6843,1277.3622) -- cycle;
    \path[draw=black,fill=white,line join=round,line cap=butt,line width=0.800pt]
      (493.9120,1123.2283) -- (649.6843,1195.8661) -- (649.6843,1277.3622) --
      (493.9120,1204.7244) -- cycle;
  \end{scope}
  \path[draw=black,line join=miter,line cap=butt,line width=0.800pt]
    (260.8804,463.1428) -- (260.8804,495.0326);
  \begin{scope}[shift={(-208.41255,-700.02081)},draw=black,fill=ccls31,line join=round]
    \path[shift={(0,308.2677)},draw=black,fill=ccls31,line join=round,line
      cap=butt,line width=0.800pt] (474.9154,884.0551) -- (474.9154,965.5512) --
      (562.2998,1006.2992) -- (562.2998,924.8031) -- cycle;
    \path[shift={(0,308.2677)},draw=black,fill=ccls31,line join=round,line
      cap=butt,line width=0.800pt] (474.9154,884.0551) -- (493.9120,875.1968) --
      (581.2965,915.9449) -- (562.2998,924.8031) -- cycle;
    \path[shift={(0,308.2677)},draw=black,fill=ccls31,line join=round,line
      cap=butt,line width=0.800pt] (562.2998,924.8031) -- (581.2965,915.9449) --
      (581.2965,997.4409) -- (562.2998,1006.2992) -- cycle;
  \end{scope}
  \path[draw=black,line join=miter,line cap=butt,line width=0.800pt]
    (336.8668,541.0956) -- (447.0472,489.7176);
  \begin{scope}[shift={(-172.24247,-709.6918)},draw=black,fill=ccls32,line join=round]
    \path[shift={(0,308.2677)},draw=black,fill=ccls32,line join=round,line
      cap=butt,line width=0.800pt] (611.6911,947.8346) -- (611.6911,1029.3307) --
      (630.6877,1038.1890) -- (630.6877,956.6929) -- cycle;
    \path[shift={(0,308.2677)},draw=black,fill=ccls32,line join=round,line
      cap=butt,line width=0.800pt] (611.6911,947.8346) -- (630.6877,938.9764) --
      (649.6843,947.8346) -- (630.6877,956.6929) -- cycle;
    \path[shift={(0,308.2677)},draw=black,fill=ccls32,line join=round,line
      cap=butt,line width=0.800pt] (630.6877,956.6929) -- (630.6877,1038.1890) --
      (649.6843,1029.3307) -- (649.6843,947.8346) -- cycle;
  \end{scope}
  \path[draw=black,line join=miter,line cap=butt,line width=0.800pt]
    (249.4824,500.3475) -- (336.8668,541.0956) -- (336.8668,622.5916);
  \path[draw=black,line join=miter,line cap=butt,line width=0.800pt]
    (416.6526,535.7806) -- (572.4249,608.4184);
  \begin{scope}[shift={(-553.24566,-46.35037)}]
    \begin{scope}[shift={(90.49286,-355.07313)},draw=black,fill=white,line join=round,line cap=butt,line width=0.800pt]
      \path[draw,fill] (227.9600,836.2200) -- (227.9600,1048.8200) --
        (383.7300,1121.4580) -- (383.7300,908.8580) -- (227.9600,836.2200) -- cycle;
      \path[draw,fill] (227.9600,836.2200) -- (338.1400,784.8400) --
        (493.9100,857.4800) -- (383.7300,908.8600) -- cycle;
      \path[shift={(0,308.2677)},draw,fill] (383.7300,813.1900) -- (493.9100,761.8120)
        -- (493.9100,549.2120) -- (383.7300,600.5900) -- cycle;
    \end{scope}
  \end{scope}
  \path[fill=black] (98.82267,555.96472) node[above right] (text3198) {\Huge $\approx$};
  \path[draw=black,dash pattern=on 4.80pt off 1.20pt,line join=miter,line
    cap=butt,miter limit=4.00,line width=1.200pt,o->] (-254.0433,480.3123) -- (-208.1249,501.5300);
  \begin{scope}[shift={(-331.70385,40.21882)},draw=black,fill=ccls5,line join=round]
    \path[shift={(-235.40819,370.25803)},draw=black,fill=ccls5,line
      join=round,line cap=butt,line width=0.800pt] (351.7989,87.9086) --
      (351.7989,105.6252) -- (370.7955,114.4834) -- (370.7955,96.7669) -- cycle;
    \path[shift={(-235.40819,370.25803)},draw=black,fill=ccls5,line
      join=round,line cap=butt,line width=0.800pt] (351.7989,87.9086) --
      (370.7955,79.0504) -- (389.7921,87.9086) -- (370.7955,96.7669) -- cycle;
    \path[shift={(-235.40819,370.25803)},draw=black,fill=ccls5,line
      join=round,line cap=butt,line width=0.800pt] (370.7955,96.7669) --
      (370.7955,114.4834) -- (389.7921,105.6252) -- (389.7921,87.9086) -- cycle;
  \end{scope}
  \path[fill=black] (-360.03314,490.31232) node[above right,align=right] (text4469) {\huge $i$-th\\ \huge entity};
  \path[draw=black,dash pattern=on 4.80pt off 1.20pt,line join=miter,line
    cap=butt,miter limit=4.00,line width=1.200pt,o->] (-197.0535,400.5879) -- (-196.3392,495.8157);
  \path[draw=black,dash pattern=on 4.80pt off 1.20pt,line join=miter,line
    cap=butt,miter limit=4.00,line width=1.200pt,o->] (-254.0433,533.4619) -- (-208.4820,512.2443);
  \path[fill=black] (-360.05345,399.72964) node[above right] (text5781) {\huge $j$-th
    entity};
  \path[fill=black] (-380.02979,585.32019) node[above right,align=right] (text5787) {\huge $k$-th\\ \huge relation};
  \path[fill=black] (50.394096,470.02493) node[above right,align=right] (text4469-2) {\huge $i$-th\\ \huge\ entity};
  \path[draw=black,dash pattern=on 4.80pt off 1.20pt,line join=miter,line
    cap=butt,miter limit=4.00,line width=1.200pt,o->] (154.3839,626.4737) -- (314.0087,551.7180);
  \path[draw=black,dash pattern=on 4.80pt off 1.20pt,line join=miter,line
    cap=butt,miter limit=4.00,line width=1.200pt,o->] (458.3298,502.4580) -- (459.0441,544.8081);
  \path[fill=black] (465.3298,498.59973) node[above right,align=left] (text5781-4) {\huge $j$-th\\ \huge entity};
  \path[fill=black] (45.397484,685.33203) node[above right,align=right] (text5787-8) {\huge $k$-th\\ \huge relation};
  \path[fill=black] (-122.16412,769.6377) node[above right] (text3077) {\Huge $\mat{Y}_k$};
  \path[fill=black] (270.37335,768.20605) node[above right] (text3148) {\Huge $\mat{E} \mat{W}_k \mat{E}^\transp$};
\end{scope}

\end{tikzpicture}
}}
  \caption{\label{fig:rescal:model} RESCAL as a
  tensor factorization of the adjacency tensor $\ten{Y}$.}
\end{figure}

RESCAL is similar to methods used in recommendation systems
\cite{koren_matrix_2009}, and to traditional tensor factorization methods
\cite{kolda_tensor_2009}. In matrix notation, \cref{eq:rescal} can be written
compactly as as ${\mat{F}_k = \mat{E} \mat{W}_k \mat{E}^\transp}$, where
$\mat{F}_k \in \R^{\Nentities \times \Nentities}$ is the matrix holding all
scores for the $k$-th relation and the $i$-th row of $\mat{E} \in \R^{\Nentities
  \times \dimentities}$ holds the latent representation of $\vec{e}_i$. See
\cref{fig:rescal:model} for an illustration. In the following, we will use this
tensor representation to derive a very efficient algorithm for parameter
estimation.

\subsubsection*{Fitting the model}
If we want to compute a probabilistic model, the parameters of RESCAL can be
estimated by minimizing the log-loss using gradient-based methods such as
stochastic gradient descent~\citep{nickel_logistic_2013}. RESCAL can also be
computed as a score-based model, which has the main advantage that we can
estimate the parameters $\Theta$ very efficiently: Due to its tensor structure
and due to the sparsity of the data, it has been shown that the RESCAL model can
be computed via a sequence of efficient closed-form updates when using the
squared-loss~\citep{nickel_three-way_2011,nickel_factorizing_2012}. In this
setting, it has been shown analytically that a single update of $\mE$ and
$\mat{W}_k$ scales linearly with the number of entities $\Nentities$, linearly
with the number of relations $\Nreln$, and linearly with the number of
\emph{observed} triples, i.e., the number of non-zero entries in
$\tY$~\citep{nickel_factorizing_2012}. We call this algorithm
RESCAL-ALS.\footnote{ALS stands for Alternating Least-Squares} In practice, a
small number (say 30 to 50) of iterated updates are often sufficient for
RESCAL-ALS to arrive at stable estimates of the parameters.
Given a current estimate of $\mE$, the updates for each $\mat{W}_k$ can be
computed in parallel to improve the scalability on knowledge graphs with a large
number of relations. Furthermore, by exploiting the special tensor structure of
RESCAL, we can derive improved updates for RESCAL-ALS that compute the estimates
for the parameters with a runtime complexity of $\BigO(\dimentities^3)$ for a
single update (as opposed to a runtime complexity of $\BigO(\dimentities^5)$ for
naive updates)~\citep{nickel_tensor_2013,chang_typed_2014}. In summary, for
relational domains that can be explained via a moderate number of latent
features, RESCAL-ALS is highly scalable and very fast to compute. For more
detail on RESCAL-ALS see also~\cref{eq:rescal-ls-loss} in \cref{sec:training}.

\subsubsection*{Decoupled Prediction}
In \cref{eq:rescal}, the probability of single relationship is computed via
simple matrix-vector products in $\BigO(\dimentities^2)$ time. Hence, once the
parameters have been estimated, the computational complexity to predict the
score of a triple depends only on the number of latent features and is
independent of the size of the graph. However, during parameter estimation, the
model can capture global dependencies due to the shared latent representations.

\subsubsection*{Relational learning results}
RESCAL has been shown to achieve state-of-the-art results on a number of
relational learning tasks. For instance, \citet{nickel_three-way_2011} showed
that RESCAL provides comparable or better relationship prediction results on a
number of small benchmark datasets compared to Markov Logic Networks (with
structure learning)~\citep{kok_statistical_2007}, the Infinite (Hidden)
Relational model~\citep{xu_infinite_2006,kemp_learning_2006}, and Bayesian
Clustered Tensor Factorization~\citep{sutskever_modelling_2009}. Moreover,
RESCAL has been used for link prediction on entire knowledge graphs such as YAGO
and DBpedia~\citep{nickel_factorizing_2012,krompas_large-scale_2014}. Aside from
link prediction, RESCAL has also successfully been applied to SRL tasks such as
entity resolution and link-based clustering. For instance, RESCAL has shown
state-of-the-art results in predicting which authors, publications, or
publication venues are likely to be identical in publication
databases~\citep{nickel_three-way_2011,nickel_tensor_2013}. Furthermore, the
semantic embedding of entities computed by RESCAL has been exploited to create
taxonomies for uncategorized data via hierarchical clusterings of entities in
the embedding space~\citep{nickel_learning_2011}.

\subsection{Other tensor factorization models}
\label{sec:related-tensor}

Various other tensor factorization methods have been explored for learning from
knowledge graphs and multi-relational data.
\citet{kolda2005tophits,franz_triplerank:_2009} factorized adjacency tensors
using the CP tensor decomposition to analyze the link structure of Web pages and
Semantic Web data respectively. \citet{drumond_predicting_2012} applied pairwise
interaction tensor factorization~\citep{rendle_pairwise_2010} to predict triples
in knowledge graphs. \citet{rendle_scaling_2013} applied factorization machines
to large uni-relational datasets in recommendation settings.
\citet{jenatton_latent_2012} proposed a tensor factorization model for knowledge
graphs with a very large number of different relations.

It is also possible to use discrete latent factors.
\citet{miettinen_boolean_2011} proposed Boolean tensor factorization to
disambiguate facts extracted with OpenIE methods and applied it to large
datasets~\citep{erdos_discovering_2013}. In contrast to previously discussed
factorizations, Boolean tensor factorizations are discrete models, where
adjacency tensors are decomposed into binary factors based on Boolean algebra.

\subsection{Matrix factorization methods}

Another approach for learning from knowledge graphs is based on matrix
factorization, where, prior to the factorization, the adjacency tensor $\tY \in
\R^{\Nentities \times \Nentities \times \Nreln}$ is reshaped into a matrix $\mY
\in \R^{\Nentities^2 \times \Nreln}$ by associating rows with subject-object
pairs $(e_i, e_j)$ and columns with relations $r_k$ (cf.\
\citep{jiang_link_2012, riedel_relation_2013}), or into a matrix $\mY \in
\R^{\Nentities \times \Nentities \Nreln}$ by associating rows with subjects
$e_i$ and columns with relation/objects $(r_k, e_j)$ (cf.\
\citep{tresp_materializing_2009,huang_scalable_2013}). Unfortunately, both of
these formulations lose information compared to tensor factorization. For
instance, if each subject-object pair is modeled via a different latent
representation, the information that the relationships $y_{ijk}$ and $y_{pjq}$
share the same object is lost. It also leads to an increased memory complexity,
since a separate latent representation is computed for each pair of entities,
requiring $\BigO(\Nentities^2 \dimentities + \Nreln \dimentities)$ parameters
(compared to $\BigO(\Nentities \dimentities + \Nreln \dimentities^2)$ parameters
for RESCAL).

\subsection{Multi-layer perceptrons}
\label{sec:MLP}

\begin{figure*}[t]
  \centering
  \vspace{-1em}
  \subfloat[RESCAL]{\resizebox{!}{2.5cm}{\label{fig:rescal:basis} \begin{tikzpicture}[scale=1,baseline,thick]
  \GraphInit[vstyle=Classic]
  \tikzset{vertex/.style =
    {draw=black,shape=circle,fill=white,minimum size=13pt,circular
      drop shadow}}
  \tikzset{split vertex/.style =
    {draw=black,shape=forbidden sign,fill=ccls31,minimum size=13pt,circular
      drop shadow,circle split part fill=ccls32}}
  \node at (-0.5,0)[vertex,fill=ccls31,label={[xshift=1.5em,yshift=-1.75em]$e_{i1}$}] (s1) {};
  \node at (0.5,0)[vertex,fill=ccls31,label={[xshift=1.5em,yshift=-1.75em]$e_{i2}$}] (s2) {};
  \node at (1.5,0)[vertex,fill=ccls31,label={[xshift=1.5em,yshift=-1.75em]$e_{i3}$}] (s3) {};
  \node at (3,0)[vertex,fill=ccls32,label={[xshift=1.5em,yshift=-1.75em]$e_{j1}$}] (o1) {};
  \node at (4,0)[vertex,fill=ccls32,label={[xshift=1.5em,yshift=-1.75em]$e_{j2}$}] (o2) {};
  \node at (5,0)[vertex,fill=ccls32,label={[xshift=1.5em,yshift=-1.75em]$e_{j3}$}] (o3) {};
  \node at (-0.75,1.25)[vertex] (s1o1) {};
  \node at (0,1.25)[vertex] (s1o2) {};
  \node at (0.75,1.25)[vertex] (s1o3) {};
  \node at (1.5,1.25)[vertex] (s2o1) {};
  \node at (2.25,1.25)[vertex] (s2o2) {};
  \node at (3,1.25)[vertex] (s2o3) {};
  \node at (3.75,1.25)[vertex] (s3o1) {};
  \node at (4.5,1.25)[vertex] (s3o2) {};
  \node at (5.25,1.25)[vertex] (s3o3) {};
  \node at (2.25,2.25)[vertex,label=above right:{$f_{ijk}$}] (out) {};

  \node at (-1.5, 0.5) (kron) {$\kron$};
  \node at (-1.5, 2) (W) {$\protect{\vw_k}$};

  \node at (-0.5,-0.5) (s1l) {};
  \node at (1.5,-0.5) (s2l) {};
  \node at (0.75,-0.775) (subl) {\footnotesize subject};
  \draw[decoration={brace,mirror},decorate] (s1l.west) -- (s2l.east);

  \node at (3,-0.5) (o1l) {};
  \node at (5,-0.5) (o2l) {};
  \node at (4,-0.775) (subl) {\footnotesize object};
  \draw[decoration={brace,mirror},decorate] (o1l.west) -- (o2l.east);

  \tikzstyle{EdgeStyle}=[->,>=stealth,thick]
  \Edge (s1)(s1o1) \Edge (s1)(s1o2) \Edge (s1)(s1o3)
  \Edge (s2)(s2o1) \Edge (s2)(s2o2) \Edge (s2)(s2o3)
  \Edge (s3)(s3o1) \Edge (s3)(s3o2) \Edge (s3)(s3o3)
  \Edge (s1o1)(out) \Edge (s1o2)(out) \Edge (s1o3)(out)
  \Edge (s2o1)(out) \Edge (s2o2)(out) \Edge (s2o3)(out)
  \Edge (s3o1)(out) \Edge (s3o2)(out) \Edge (s3o3)(out)
  \Edge (o1)(s1o1) \Edge (o1)(s2o1) \Edge (o1)(s3o1)
  \Edge (o2)(s1o2) \Edge (o2)(s2o2) \Edge (o2)(s3o2)
  \Edge (o3)(s1o3) \Edge (o3)(s2o3) \Edge (o3)(s3o3)
\end{tikzpicture}

}}
  \hspace{3em}
  \subfloat[ER-MLP]{\label{fig:kv-basis}\resizebox{!}{2.5cm}{\begin{tikzpicture}[scale=1,baseline,thick]
  \GraphInit[vstyle=Classic]
  \tikzset{vertex/.style =
    {draw=black,shape=circle,fill=white,minimum size=13pt,circular
      drop shadow}}

  \node at (0,0)[vertex,fill=ccls31,label={[xshift=1.5em,yshift=-1.75em]$e_{i1}$}] (u1) {};
  \node at (1,0)[vertex,fill=ccls31,label={[xshift=1.5em,yshift=-1.75em]$e_{i2}$}] (u2) {};
  \node at (2,0)[vertex,fill=ccls31,label={[xshift=1.5em,yshift=-1.75em]$e_{i3}$}] (u3) {};
  \node at (3,0)[vertex,fill=ccls32,label={[xshift=1.5em,yshift=-1.75em]$e_{k1}$}] (w1) {};
  \node at (4,0)[vertex,fill=ccls32,label={[xshift=1.5em,yshift=-1.75em]$e_{k2}$}] (w2) {};
  \node at (5,0)[vertex,fill=ccls32,label={[xshift=1.5em,yshift=-1.75em]$e_{k3}$}] (w3) {};
  \node at (6,0)[vertex,fill=ccls33,label={[xshift=1.5em,yshift=-1.75em]$r_{j1}$}] (v1) {};
  \node at (7,0)[vertex,fill=ccls33,label={[xshift=1.5em,yshift=-1.75em]$r_{j2}$}] (v2) {};
  \node at (8,0)[vertex,fill=ccls33,label={[xshift=1.5em,yshift=-1.75em]$r_{j3}$}] (v3) {};
  \node at (1,1.5)[vertex,label=right:$h_{c1}$] (b1) {$g$};
  \node at (4,1.5)[vertex,label=right:$h_{c2}$] (b2) {$g$};
  \node at (7,1.5)[vertex,label=right:$h_{c3}$] (b3) {$g$};
  \node at (4,2.5)[vertex,label=above right:{$f_{ijk}$}] (out) {};

  \node at (-0.5, 1) (W) {$C$};
  \node at (-0.5, 2.25) (beta) {$\vw$};

  \node at (0,-0.5) (u1l) {};
  \node at (2,-0.5) (u2l) {};
  \node at (1,-0.775) (subl) {\footnotesize subject};
  \draw[decoration={brace,mirror},decorate] (u1l.west) -- (u2l.east);

  \node at (3,-0.5) (w1l) {};
  \node at (5,-0.5) (w2l) {};
  \node at (4,-0.775) (subl) {\footnotesize object};
  \draw[decoration={brace,mirror},decorate] (w1l.west) -- (w2l.east);

  \node at (6,-0.5) (v1l) {};
  \node at (8,-0.5) (v2l) {};
  \node at (7,-0.775) (subl) {\footnotesize predicate};
  \draw[decoration={brace,mirror},decorate] (v1l.west) -- (v2l.east);
  \tikzstyle{EdgeStyle}=[->,>=stealth,thick]
  \Edge (u1)(b1) \Edge (u1)(b2) \Edge (u1)(b3)
  \Edge (u2)(b1) \Edge (u2)(b2) \Edge (u2)(b3)
  \Edge (u3)(b1) \Edge (u3)(b2) \Edge (u3)(b3)
  \Edge (w1)(b1) \Edge (w1)(b2) \Edge (w1)(b3)
  \Edge (w2)(b1) \Edge (w2)(b2) \Edge (w2)(b3)
  \Edge (w3)(b1) \Edge (w3)(b2) \Edge (w3)(b3)
  \Edge (v1)(b1) \Edge (v1)(b2) \Edge (v1)(b3)
  \Edge (v2)(b1) \Edge (v2)(b2) \Edge (v2)(b3)
  \Edge (v3)(b1) \Edge (v3)(b2) \Edge (v3)(b3)
  \Edge (b1)(out) \Edge (b2)(out) \Edge (b3)(out)

\end{tikzpicture}

}}
  \caption{Visualization of RESCAL and the ER-MLP model as Neural Networks.
    Here, $\dimentities = \dimreln = 3$ and $\dimA=3$. Note, that the inputs are
    latent features. The symbol $g$ denotes the application of the function
    $g(\cdot)$.}
\end{figure*}
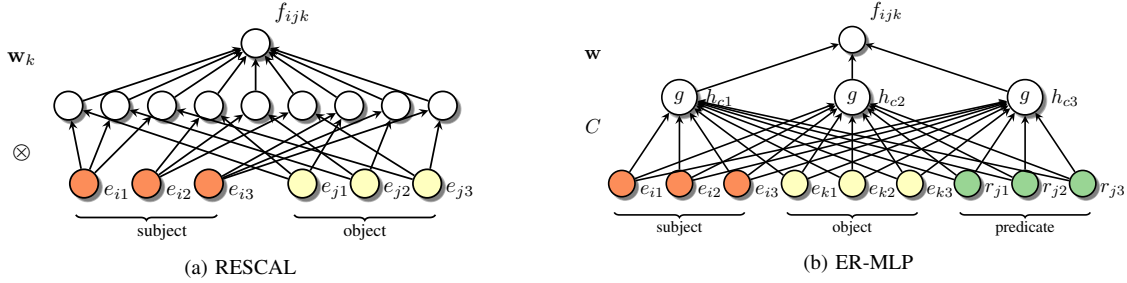

We can interpret RESCAL as creating composite representations of triples and
predicting their existence from this representation. In particular, we can
rewrite RESCAL as
\begin{align}
f_{ijk}^{\text{RESCAL}}  & \mdef \vw^\transp_k \bm{\phi}_{ij}^{\text{RESCAL}} \label{eq:rescal:fun}\\
\bm{\phi}_{ij}^{\text{RESCAL}} & \mdef \ve_j \kron \ve_i \label{eq:rescal:rep},
\end{align}
where $\vw_k = \vect{\mat{W}_k}$. \Cref{eq:rescal:fun} follows from
\cref{eq:rescal} via the equality ${\vect{\mat{AXB}} = (\mat{B}^\transp \kron
  \mat{A})\vect{\mat{X}}}$. Hence, RESCAL represents pairs of entities $(e_i,
e_j)$ via the tensor product of their latent feature representations
(\cref{eq:rescal:rep}) and predicts the existence of the triple $x_{ijk}$ from
$\vphi_{ij}$ via $\vw_k$ (\cref{eq:rescal:fun}). See also
\cref{fig:rescal:basis}. For a further discussion of the tensor product to
create composite latent representations please
see~\citet{smolensky_tensor_1990,halford_processing_1998,plate_common_1997}.

Since the tensor product explicitly models \emph{all pairwise} interactions,
RESCAL can require a lot of parameters when the number of latent features are
large (each matrix $\mat{W}_k$ has $\dimentities^2$ entries). This can, for
instance, lead to scalability problems on knowledge graphs with a large number
of relations.

In the following we will discuss models based on multi-layer perceptrons (MLPs),
also known as feedforward neural networks. In the context of multidimensional
data they can be referred to a muliway neural networks. This approach allows us
to consider alternative ways to create composite triple representations and to
use nonlinear functions to predict their existence.

In particular, let us define the following E-MLP model (E for entity):
\begin{align}
  f_{ijk}^{\text{E-MLP}} & \mdef \vw_k^\transp \vg(\vh_{ijk}^a) \label{eqn:EMLP} \\
  \vh^a_{ijk} & \mdef \mA_k^\transp \vphi_{ij}^{\text{E-MLP}} \label{eqn:EMLP:basis}\\
  \vphi_{ij}^{\text{E-MLP}} & \mdef [\ve_i; \ve_j]
\end{align}
where $\vg(\vu)=[g(u_1), g(u_2), \ldots] $ is the function $g$ applied
element-wise to vector $\vu$; one often uses the nonlinear function $g(u) =
\tanh(u)$.

Here $\vh^a$ is an additive hidden layer, which is deriving by adding together
different weighed components of the entity representations. In particular, we
create a composite representation $\vphi_{ij}^{\text{E-MLP}} = [\ve_i; \ve_j]
\in \R^{2\dimA}$ via the \emph{concatenation} of $\ve_i$ and $\ve_j$. However,
concatenation alone does not consider any interactions between the latent
features of $e_i$ and $e_j$. For this reason, we add a (vector-valued) hidden
layer $\vh_a$ of size $\dimA$, from which the final prediction is derived via
$\vw_k^\transp \vg(\vh_a)$. The important difference to tensor-product models
like RESCAL is that we \emph{learn} the interactions of latent features via the
matrix $\mA_k$ (\cref{eqn:EMLP:basis}), while the tensor product considers
always all possible interactions between latent features. This adaptive approach
can reduce the number of required parameters significantly, especially on
datasets with a large number of relations.

One disadvantage of the E-MLP is that it has to define a vector $\vw_k$ and a
matrix $\mA_k$ for every possible relation, which requires $H_a + (H_a \times
2H_e)$ parameters per relation. An alternative is to embed the relation itself,
using a $\dimreln$-dimensional vector $\vr_k$. We can then define
\begin{align}
f_{ijk}^{\text{ER-MLP}} & \mdef \vw^\transp \vg(\vh_{ijk}^c)\\
\vh_{ijk}^c & \mdef \mC^\transp \vphi_{ijk}^{\text{ER-MLP}}\\
\vphi_{ijk}^{\text{ER-MLP}} & \mdef [ \ve_i; \ve_j; \vr_k] .
\end{align}
We call this model the ER-MLP, since it applies an MLP to an embedding of the
entities and relations. Please note that ER-MLP uses a global weight vector for
all relations. This model was used in the KV project (see Section~\ref{sec:kv}),
since it has many fewer parameters than the E-MLP (see Table~\ref{tab:embed});
the reason is that $\mC$ is independent of the relation $k$.

It has been shown in \citep{mikolov_efficient_2013} that MLPs can learn to put
``semantically similar'' words close by in the embedding space, even if they are
not explicitly trained to do so. In \cite{dong_knowledge_2014}, they show a
similar result for the semantic embedding of relations using ER-MLP. For
example, \Cref{tab:kv-mlp:nn} shows the nearest neighbors of latent
representations of selected relations that have been computed with a 60
dimensional model on Freebase. Numbers in parentheses represent squared
Euclidean distances. It can be seen that ER-MLP puts semantically related
relations near each other. For instance, the closest relations to the
\textit{children} relation are \textit{parents}, \textit{spouse}, and
\textit{birthplace}.

\begin{table}
  \caption{Semantic Embeddings of KV-MLP on Freebase}
\label{tab:kv-mlp:nn}
  \centering
  \begin{tabular}{lllllll}
    \toprule
    \textbf{Relation} & \multicolumn{6}{c}{\textbf{Nearest Neighbors}} \\
    \cmidrule(r){1-1} \cmidrule(l){2-7}
    children & parents & \hspace{-.75em}(0.4) & spouse & \hspace{-.75em}(0.5) & birth-place & \hspace{-.75em}(0.8) \\
    birth-date & children & \hspace{-.75em}(1.24) & gender & \hspace{-.75em}(1.25) & parents &
                                                                 \hspace{-.75em}(1.29) \\
    edu-end\tablefootnote{The relations
\textit{edu-start}, \textit{edu-end}, \textit{job-start},
\textit{job-end} represent the start and end dates of
attending an educational institution and holding a
particular job, respectively} & job-start & \hspace{-.75em}(1.41) & edu-start & \hspace{-.75em}(1.61) & job-end & \hspace{-.75em}(1.74) \\
    \bottomrule
  \end{tabular}
\end{table}

\subsection{Neural tensor networks}
\label{sec:otherNN}
\label{sec:NTN}

\begin{table*}[t]
  \caption{Summary of the latent feature models.
    $\vh_a$, $\vh_b$ and $\vh_c$ are hidden layers of the neural network; see text for details.}
  \label{tab:embed}
  \centering
  \begin{tabular}{llllll}
    \toprule
    \textbf{Method} & $f_{ijk}$ & \ $\mA_k$ & $\mC$ & $\tB_k$ & \textbf{Num. Parameters}\\
    \cmidrule(r){1-1}\cmidrule(lr){2-2}\cmidrule(lr){3-3}\cmidrule(lr){4-4}\cmidrule(lr){5-5}\cmidrule(l){6-6}
    RESCAL \cite{nickel_factorizing_2012} 
                    & $\vw_k^\transp \vh_{ijk}^b$
                    & - 
                    & - 
                    & $[\delta_{1,1},\ldots,\delta_{H_e,H_e}]$
                    & $\Nreln H_e^2 + \Nentities H_e$ \\
    E-MLP \cite{socher_reasoning_2013} 
                    & $\vw_k^\transp \vg(\vh_{ijk}^a)$
                    & $[\mA_k^s;\phantom{-}\mA_k^o]$ 
                    & - 
                    & - 
                    & $\Nreln (H_a + H_a \times 2 H_e) + \Nentities H_e$ \\
    ER-MLP \cite{dong_knowledge_2014} 
                    & $\vw^\transp \vg(\vh_{ijk}^c)$
                    & - 
                    & $\mC$ 
                    & - 
                    & $H_c + H_c \times (2 H_e + H_r) + \Nreln H_r + \Nentities H_e$ \\
    NTN \cite{socher_reasoning_2013} 
                    & $\vw_k^\transp \vg([\vh_{ijk}^a; \vh_{ijk}^b])$
                    & $[\mA_k^s; \mA_k^o]$ 
                    & - 
                    & $[\mB_k^1, \ldots, \mB_k^{H_b}]$ 
                    & $\Nentities^2 H_b + \Nreln(H_b + H_a) +2 \Nreln H_e H_a + \Nentities H_e$ \\
    Structured Embeddings \cite{bordes_learning_2011}  
                    & $-\|\vh_{ijk}^a\|_1$
                    & $[\mA_k^s; -\mA_k^o]$  
                    & - 
                    & - 
                    & $2 \Nreln H_e H_a + \Nentities H_e$ \\
    TransE \cite{bordes_translating_2013} 
                    & $-(2h_{ijk}^a - 2h_{ijk}^b + \|\vr_k\|_2^2)$
                    & $[\vr_k; -\vr_k]$ 
                    & - 
                    & $\mI$ 
                    & $\Nreln H_e  + \Nentities H_e$ \\
    \bottomrule
  \end{tabular}
\end{table*}

We can combine traditional MLPs with bilinear models, resulting in what
\cite{socher_reasoning_2013} calls a ``neural tensor network'' (NTN). More
precisely, we can define the NTN model as follows:
\begin{align}
f_{ijk}^{\text{NTN}} & \mdef \vw_k^\transp \vg([\vh_{ijk}^a; \vh_{ijk}^b])
\label{eqn:fNTN} \\
\vh_{ijk}^a &\mdef \mA_k^\transp [\ve_i; \ve_j] \\
\vh_{ijk}^b & \mdef \left[\ve_i^\transp \mB_k^1 \ve_j ,
\ldots,\ve_i^\transp \mB_k^{H_b} \ve_j \right]
\end{align}
Here $\tB_k$ is a tensor, where the $\ell$-th slice $\mB_k^\ell$ has size $H_e
\times H_e$, and there are $H_b$ slices. We call $\vh_{ijk}^b$ a bilinear hidden
layer, since it is derived from a weighted combination of multiplicative terms.

NTN is a generalization of the RESCAL approach, as we explain in
Section~\ref{sec:rescalAsNTN}. Also, it uses the additive layer from the E-MLP
model. However, it has many more parameters than the E-MLP or RESCAL models.
Indeed, the results in \cite{yang_embedding_2014} and \cite{dong_knowledge_2014}
both show that it tends to overfit, at least on the (relatively small) datasets
uses in those papers.

\subsection{Latent distance models}

Another class of models are latent distance models (also known as latent space
models in social network analysis), which derive the probability of
relationships from the \emph{distance} between latent representations of
entities: entities are likely to be in a relationship if their latent
representations are close according to some distance measure. For uni-relational
data, \citet{hoff_latent_2002} proposed this approach first in the context of
social networks by modeling the probability of a relationship $x_{ij}$ via the
score function $f(e_i, e_j) = -d(\ve_i,\ve_j)$ where $d(\cdot,\cdot)$ refers to
an arbitrary distance measure such as the Euclidean distance.

The structured embedding (SE) model~\citep{bordes_learning_2011} extends this
idea to multi-relational data by modeling the score of a triple $x_{ijk}$ as:
\begin{equation}
  \label{eq:se}
  f_{ijk}^{\text{SE}} \mdef -\|\mA_k^s\ve_i - \mA_k^o\ve_j\|_1 = -\|\vh_{ijk}^a\|_1
\end{equation}
where $\mA_k = [\mA_k^s; -\mA_k^o]$. In \cref{eq:se} the matrices $\mA_k^s$,
$\mA_k^o$ transform the global latent feature representations of entities to
model relationships specifically for the $k$-th relation. The transformations
are learned using the ranking loss in a way such that pairs of entities in
existing relationships are closer to each other than entities in non-existing
relationships.

To reduce the number of parameters over the SE model, the TransE
model~\citep{bordes_translating_2013} translates the latent feature
representations via a relation-specific offset instead of transforming them via
matrix multiplications. In particular, the score of a triple $x_{ijk}$ is
defined as:
\begin{equation}
  \label{eq:transe}
  f_{ijk}^{\text{TransE}} \mdef -d(\ve_i + \vr_k, \ve_j)  .
\end{equation}
This model is inspired by the results in \cite{mikolov_efficient_2013}, who
showed that some relationships between words could be computed by their vector
difference in the embedding space. As noted in \cite{yang_embedding_2014}, under
unit-norm constraints on $\ve_i,\ve_j$ and using the squared Euclidean distance,
we can rewrite \cref{eq:transe} as follows:
\begin{equation}
  f_{ijk}^{\text{TransE}}
  = -(2 \vr_k^\transp(\ve_i - \ve_j) - 2 \ve_i^\transp \ve_j + \|\vr_k\|^2_2)
\end{equation}
Furthermore, if we assume $\mA_k = [\vr_k; -\vr_k]$, so that ${h_{ijk}^a = [\vr_k;
-\vr_k]^T [\ve_i; \ve_j] = \vr_k^T(\ve_i - \ve_j)}$, and $\tB_k=\mI$, so that
${h_{ijk}^b = \ve_i^T \ve_j}$, then we can rewrite this model as follows:
\begin{equation}
  f_{ijk}^{\text{TransE}} = -(2h_{ijk}^a -2h_{ijk}^b + \|\vr_k\|_2^2) .
\end{equation}

\subsection{Comparison of models}
Table~\ref{tab:embed} summarizes the different models we have discussed. A
natural question is: which model is best? \cite{dong_knowledge_2014} showed that
the ER-MLP model outperformed the NTN model on their particular dataset.
\cite{yang_embedding_2014} performed more extensive experimental comparison of
these models, and found that RESCAL (called the bilinear model) worked best on two
link prediction tasks. However, clearly the best model will be dataset
dependent.


\section{Graph Feature Models}
\label{sec:observable}

In this section, we assume that the existence of an edge can be predicted by
extracting features from the observed edges in the graph. For example, due to
social conventions, parents of a person are often married, so we could predict
the triple \textit{(John, marriedTo, Mary)} from the existence of the path
\mbox{\textit{John} \praright{parentOf} \textit{Anne} \praleft{parentOf}
  \textit{Mary}}, representing a common child. In contrast to latent feature
models, this kind of reasoning explains triples directly from the observed
triples in the knowledge graph. We will now discuss some models of this kind.

\subsection{Similarity measures for uni-relational data}
\label{sec:social_networks}

Observable graph feature models are widely used for link prediction in graphs that consist
only of a single relation, e.g.,  social network analysis
(friendships between people), biology (interactions of proteins), and
Web mining (hyperlinks between Web sites).  The intuition
behind these methods is that similar entities are likely to be related
(homophily) and that the similarity of entities can be derived from
the neighborhood of nodes or from the existence of paths between
nodes.
For this purpose, various indices have been proposed to measure
the similarity of entities, which can be classified into local, global,
and quasi-local approaches~\citep{lu_link_2011}.

Local similarity
indices such as \emph{Common Neighbors}, the \emph{Adamic-Adar}
index~\citep{adamic_friends_2003} or \emph{Preferential
  Attachment}~\citep{barabasi_emergence_1999} derive the similarity of
entities from their number of common neighbors or their absolute
number of neighbors. Local similarity indices are fast to compute for
single relationships and scale well to large knowledge graphs as their
computation depends only on the direct neighborhood of the involved
entities. However, they can be too localized to capture important
patterns in relational data and cannot model long-range or global
dependencies.

Global similarity indices such as the \emph{Katz}
index~\citep{katz_new_1953} and the \emph{Leicht-Holme-Newman}
index~\citep{leicht_vertex_2006} derive the similarity of entities
from the ensemble of all paths between entities, while indices like
\emph{Hitting Time}, \emph{Commute Time}, and
\emph{PageRank}~\citep{brin_anatomy_1998} derive the similarity of
entities from random walks on the graph.  Global similarity indices
often provide  significantly better predictions than local indices, but
are also computationally more expensive~\citep{lu_link_2011,liben-nowell_link-prediction_2007}.

Quasi-local similarity indices like the \emph{Local Katz}
index~\citep{liben-nowell_link-prediction_2007} or \emph{Local Random
  Walks}~\citep{liu_link_2010} try to balance predictive accuracy and
computational complexity by deriving the similarity of entities from
paths and random walks of \emph{bounded length}.

In \cref{sec:pra}, we will discuss an approach that extends this idea
of quasi-local similarity indices for uni-relational networks to learn
from large multi-relational knowledge graphs.


\subsection{Rule Mining and Inductive Logic Programming}
Another class of models that works on the observed variables of a knowledge
graph extracts rules via mining methods and uses these extracted rules to infer
new links. The extracted rules can also be used as a basis for Markov Logic as discussed in
\cref{sec:MRF}.
For instance, ALEPH is an Inductive Logic Programming (ILP) system that attempts
to learn rules from relational data via inverse
entailment~\citep{muggleton1995inverse} (For more information on ILP see e.g.,
\citep{muggleton_inductive_1991,de_raedt_logical_2008,quinlan_learning_1990}).
AMIE is a rule mining system that extracts logical rules (in particular Horn
clauses) based on their support in a knowledge
graph~\citep{galarraga_amie:_2013,galarraga2015amie}. In contrast to ALEPH,
AMIE can handle the open-world assumption of knowledge graphs and has shown to
be up to three orders of magnitude faster on large knowledge
graphs~\citep{galarraga2015amie}. The basis for the Semantic Web is Description
Logic and \citep{lisi2010inductive,Amato06reasoning,lehmann2009dllearner}
describe approaches for logic-oriented machine learning approaches in this
context.
Also to mention are data mining
approaches for knowledge graphs as described
in~\citep{rettinger2012mining,losch2012kernels,minervini2014learning}.
An advantage of rule-based systems is that they are easily interpretable as the
model is given as a set of logial rules. However, rules over observed variables cover usually
only a subset of patterns in knowledge graphs (or relational data) and useful
rules can be \mbox{challenging to learn}.



\subsection{Path Ranking Algorithm}
\label{sec:pra}

The Path Ranking Algorithm (PRA)~\citep{lao_relational_2010,lao_random_2011}
extends the idea of using random walks of bounded lengths for predicting links
in multi-relational knowledge graphs. In particular, let $\pi_{L}(i,j,k,t)$
denote a path of length $L$ of the form $e_i \stackrel{r_1}{\rightarrow} e_2
\stackrel{r_2}{\rightarrow} e_3 \cdots \stackrel{r_L}{\rightarrow} e_j$, where
$t$ represents the sequence of edge types $t=(r_1,r_2,\ldots,r_L)$. We also
require there to be a direct arc $e_i \stackrel{r_k}{\rightarrow} e_j$,
representing the existence of a relationship of type $k$ from $e_i$ to $e_j$.
Let $\Pi_{L}(i,j,k)$ represent the set of all such paths of length $L$, ranging
over path types $t$. (We can discover such paths by enumerating all
(type-consistent) paths from entities of type $e_i$ to entities of type $e_j$.
If there are too many relations to make this feasible, we can perform random
sampling.)

We can compute the probability of following such a path by assuming that at each
step, we follow an outgoing link uniformly at random. Let
$\prob(\pi_{L}(i,j,k,t))$ be the probability of this particular path; this can
be computed recursively by a sampling procedure, similar to PageRank (see
\cite{lao_random_2011} for details). The key idea in PRA is to use these path
probabilities as features for predicting the probability of missing edges. More
precisely, define the feature vector
\begin{equation}
  \vphi_{ijk}^{\text{PRA}} = [\prob(\pi): \pi \in \Pi_{L}(i,j,k)]
\end{equation}
We can then predict the edge probabilities using logistic regression:
\begin{align}
  f_{ijk}^{\text{PRA}} &\mdef \vw_k^\transp \vphi_{ijk}^{\text{PRA}}
\end{align}

\subsubsection*{Interpretability}
\label{sec:PRArules}
A useful property of PRA is that its model is easily interpretable. In
particular, relation paths can be regarded as bodies of weighted rules --- more
precisely Horn clauses --- where the weight specifies how predictive the body of
the rule is for the head. For instance, \cref{tab:pra:path-examples} shows some
relation paths along with their weights that have been learned by PRA in the KV
project (see Section~\ref{sec:kv}) to predict which college a person attended,
i.e., to predict triples of the form \textit{(p, college, c)}. The first
relation path in \cref{tab:pra:path-examples} can be interpreted as follows:
\emph{it is likely that a person attended a college if the sports team that
  drafted the person is from the same college}. This can be written in the form
of a Horn clause as follows:
\begin{rulelist}
  (p, college, c) $\gets$ (p, draftedBy, t) $\land$ (t, school, c) .
\end{rulelist}
By using a sparsity promoting prior on $\vw_k$, we can perform
feature selection, which is equivalent to rule learning.


\subsubsection*{Relational learning results}

PRA has been shown to outperform the ILP method
FOIL~\citep{quinlan_learning_1990} for link prediction in
NELL~\citep{lao_random_2011}. It has also been shown to have comparable
performance to ER-MLP on link prediction in KV: PRA obtained a result of 0.884
for the area under the ROC curve, as compared to 0.882 for
ER-MLP~\citep{dong_knowledge_2014}.

\begin{table}
  \caption{Examples of paths learned by PRA on Freebase to predict
  which college a person attended}
  \label{tab:pra:path-examples}
  \centering
 \hspace*{-1.2em}
  \begin{tabular}{lcccc}
  \toprule
  \textbf{Relation Path} & \textbf{F1} & \textbf{Prec} & \textbf{Rec} & \textbf{Weight} \\
  \cmidrule(l){1-1}\cmidrule(lr){2-2} \cmidrule(lr){3-3}
    \cmidrule(lr){4-4} \cmidrule(r){5-5}
  \textit{({draftedBy}, {school})} & 0.03 & 1.0 & 0.01 & 2.62 \\
  \textit{({sibling(s)}, {sibling}, {education}, {institution})} & 0.05  & 0.55 & 0.02 & 1.88 \\
  \textit{({spouse(s)}, {spouse}, {education}, {institution})} & 0.06  & 0.41 & 0.02 & 1.87 \\
  \textit{({parents}, {education}, {institution})} & 0.04  & 0.29 & 0.02 & 1.37 \\
  \textit{({children}, {education}, {institution})} & 0.05  & 0.21 & 0.02 & 1.85 \\
  \textit{({placeOfBirth}, {peopleBornHere}, {education})} & 0.13  & 0.1 & 0.38 & 6.4 \\
  \textit{({type}, {instance}, {education}, {institution})} & 0.05  & 0.04 & 0.34 & 1.74 \\
  \textit{({profession}, {peopleWithProf.}, {edu.}, {inst.})} & 0.04  & 0.03 & 0.33 & 2.19\\
  \bottomrule
  \end{tabular}
\end{table}


\section{Combining latent and graph feature models}
\label{sec:combining}

It has been observed experimentally (see, e.g., \cite{dong_knowledge_2014}) that
neither state-of-the-art relational latent feature models (RLFMs) nor
state-of-the-art graph feature models are superior for learning from knowledge
graphs. Instead, the strengths of latent and graph-based models are often
complementary (see e.g., \cite{Toutanova2015}), as both families focus
on different aspects of relational data:
\begin{itemize}
\item Latent feature models are well-suited for modeling global relational
  patterns via newly introduced latent variables. They are computationally
  efficient if triples can be explained with a small number of latent variables.
\item Graph feature models are well-suited for modeling local and quasi-local
  graphs patterns. They are computationally efficient if triples can be
  explained from the neighborhood of entities or from short paths in the graph.
\end{itemize}

There has also been some theoretical work comparing these two approaches
\cite{nickel_reducing_2014}. In particular, it has been shown that tensor
factorization can be inefficient when relational data consists of a large number
of strongly connected components. Fortunately, such ``problematic'' relations
can often be handled efficiently via graph-based models. A good example is the
\textit{marriedTo} relation: One marriage corresponds to a single strongly
connected component, so data with a large number of marriages would be difficult
to model with RLFMs. However, predicting \textit{marriedTo} links via
graph-based models is easy: the existence of the triple \textit{(John,
  marriedTo, Mary)} can be simply predicted from the existence of \textit{(Mary,
  marriedTo, John)}, by exploiting the symmetry of the relation. If the
\textit{(Mary, marriedTo, John)} edge is unknown, we can use statistical
patterns, such as the existence of shared children.

Combining the strengths of latent and graph-based models is therefore
a promising approach to increase the predictive performance of graph
models. It typically also speeds up the training. We now discuss some ways
of combining these two kinds of models.

\subsection{Additive relational effects model}
\label{sec:are}

\citep{nickel_reducing_2014} proposed the \emph{additive relational effects}
(ARE), which is a way to combine RLFMs with observable graph models. In
particular, if we combine RESCAL with PRA, we get
\begin{equation}
  \label{eq:model-are}
  f_{ijk}^{\text{RESCAL+PRA}} =
  \vw_k^{(1)\transp} \vphi^{\text{RESCAL}}_{ij} +
  \vw_k^{(2)\transp} \vphi^{\text{PRA}}_{ijk} .
\end{equation}
ARE models can be trained by alternately optimizing the RESCAL parameters with
the PRA parameters. The key benefit is now RESCAL only has to model the
``residual errors'' that cannot be modelled by the observable graph patterns.
This allows the method to use much lower latent dimensionality, which
significantly speeds up training time. The resulting combined model also has
increased accuracy \citep{nickel_reducing_2014}.

\subsection{Other combined models}
\label{sec:other-combining}

In addition to ARE, further models have been explored to learn jointly from
latent and observable patterns on relational data.
\citet{jiang_link_2012,riedel_relation_2013} combined a latent feature model
with an additive term to learn from latent and neighborhood-based information on
multi-relational data, as follows:\footnote{
\citet{riedel_relation_2013} 
considered an additional term
$ f_{ijk}^{\text{UNI}} \mdef f_{ijk}^{\text{ADD}} + \vw_k^\transp \vphi_{ij}^{\text{SUB+OBJ}}$,
where $\vphi_{ij}^{\text{SUB+OBJ}}$ is a (non-composite) latent feature
representation of subject-object pairs.
}
\begin{align}
  f_{ijk}^{\text{ADD}} & \mdef
                         \vw_{k, j}^{(1)\transp}\vphi^{\text{SUB}}_{i} +
                         \vw_{k, i}^{(2)\transp}\vphi^{\text{OBJ}}_{j} +
                         \vw_k^{(3)\transp}\vphi^{\text{N}}_{ijk} \\
  \vphi_{ijk}^{\text{N}}& \mdef [y_{ijk'} : k^\prime \neq k] \label{eq:addschema-neighbor}
\end{align}
Here, $\vphi^{\text{SUB}}_{i}$ is the latent representation of entity $e_i$ as a
subject and $\vphi^{\text{OBJ}}_{j}$ is the latent representation of entity
$e_j$ as an object. The term $\vphi_{ijk}^\text{N}$ captures patterns
efficiently where the existence of a triple $y_{ijk'}$ is predictive of another
triple $y_{ijk}$ between the same pair of entities (but of a different relation
type). For instance, if Leonard Nimoy was \textit{born in} Boston, it is also
likely that he \textit{lived in} Boston. This dependency between the relation
types \textit{bornIn} and \textit{livedIn} can be modeled in
\cref{eq:addschema-neighbor} by assigning a large weight to
$w_{\text{bornIn},\text{livedIn}}$.

ARE and the models of \citet{jiang_link_2012} and \citet{riedel_relation_2013}
are similar in spirit to the model of \citet{koren_factorization_2008}, which
augments SVD (i.e., matrix factorization) of a rating matrix with additive terms
to include local neighborhood information. Similarly, factorization
machines~\citep{rendle_factorization_2012} allow to combine latent and
observable patterns, by modeling higher-order interactions between input
variables via low-rank factorizations \cite{drumond_predicting_2012}.

An alternative way to combine different prediction systems is to fit them
separately, and use their outputs as inputs to another ``fusion'' system. This
is called stacking~\citep{wolpert_stacked_1992}. For instance,
\citep{dong_knowledge_2014} used the output of PRA and ER-MLP as scalar
features, and learned a final ``fusion'' layer by training a binary classifier.
Stacking has the advantage that it is very flexible in the kinds of models that
can be combined. However, it has the disadvantage that the individual models
cannot cooperate, and thus any individual model needs to be more complex than in
a combined model which is trained jointly. For example, if we fit RESCAL
separately from PRA, we will need a larger number of latent features than if we
fit them jointly.


\section{Training SRL models on knowledge graphs}
\label{sec:training}
In this section we discuss aspects of training the previously discussed models
that are specific to knowledge graphs, such as how to handle the open-world
assumption of knowledge graphs, how to exploit sparsity, and how to perform
model selection.

\subsection{Penalized maximum likelihood training}
Let us assume we have a set of $\Nsamples$ observed triples and let the $n$-th
triple be denoted by $x^n$. Each observed triple is either true (denoted $y^n =
1$) or false (denoted $y^n = 0$). Let this labeled dataset be denoted by $\calD
= \setdef{(x^n, y^n)}{n=1,\ldots,\Nsamples}$. Given this, a natural way to
estimate the parameters $\Theta$ is to compute the maximum \emph{a posteriori}
(MAP) estimate:
\begin{equation}
  \max_{\Theta} \sum_{n=1}^{\Nsamples} \log
  \operatorname{Ber}(y^n \,|\, \sigma(f(x^n; \Theta))) + \log p(\Theta\,|\,\lambda)
\end{equation}
where $\lambda$ controls the strength of the prior. (If the prior is uniform,
this is equivalent to maximum likelihood training.) We can equivalently state
this as a regularized loss minimization problem:
\begin{equation}
  \min_{\Theta} \sum_{n=1}^N \loss( \sigma(f(x^n; \Theta)), y^n) + \lambda \operatorname{reg}(\Theta)
\end{equation}
where $\loss(p, y)=-\log \operatorname{Ber}(y|p)$ is the log loss function.
Another possible loss function is the squared loss, $\loss(p,y) = (p-y)^2$.
Using the squared loss can be especially efficient in combination with a
closed-world assumption (CWA). For instance, using the squared loss and the CWA,
the minimization problem for RESCAL becomes
\begin{equation}
  \label{eq:rescal-ls-loss}
  \min_{\mE,  \{\mW_k\} } \sum_k \|\mY_k - \mE \mW_k \mE^\transp \|_F^2 + \lambda_1
  \|\mE\|_F^2 + \lambda_2 \sum_k \|\mW_k\|_F^2  .
\end{equation}
where $\lambda_1,\lambda_2 \geq 0$ control the degree of regularization. The
main advantage of \cref{eq:rescal-ls-loss} is that it can be optimized via
RESCAL-ALS, which consists of a sequence of very efficient, closed-form updates
whose computational complexity depends only on the non-zero entries in
$\ten{Y}$~\citep{nickel_three-way_2011,nickel_factorizing_2012}.
We discuss some other loss functions below.

\subsection{Where do the negative examples come from?}
\label{sec:LCWA}
One important question is where the labels $y^n$ come from. The problem is that
most knowledge graphs only contain positive training examples, since, usually,
they do not encode false facts. Hence $y^n=1$ for all $(x^n, y^n) \in \calD$. To
emphasize this, we shall use the notation $\calD^+$ to represent the observed
positive (true) triples: $\calD^+ = \setdef{x^n \in \calD}{y^n = 1}$. Training
on all-positive data is tricky, because the model might easily over generalize.

One way around this is as to make a closed world assumption and assume that
all (type consistent) triples that are not in $\calD^+$ are false.
We will denote this negative set as $\calD^- = \setdef{x^n \in \calD}{y^n
  = 0}$. However, for incomplete knowledge graphs this assumption will be
  violated. Moreover, $\calD^-$ might be very large, since the number of
false facts is much larger than the number of true facts. This can lead to
scalability issues in training methods that have to consider all negative examples.

An alternative approach to generate negative examples is to exploit known
constraints on the structure of a knowledge graph: Type constraints for
predicates (persons are only married to persons), valid value ranges for
attributes (the height of humans is below 3 meters), or functional constraints
such as mutual exclusion (a person is born exactly in one city) can all be used
for this purpose. Since such examples are based on the violation of hard
constraints, it is certain that they are indeed negative examples.
Unfortunately, functional constraints are scarce and negative examples based on
type constraints and valid value ranges are usually not sufficient to train
useful models: While it is relatively easy to predict that a person is married
to another person, it is difficult to predict to which person in particular. For
the latter, examples based on type constraints alone are not very informative. A
better way to generate negative examples is to ``perturb'' true triples. In
particular, let us define
\begin{align*}
  \calD^- & = \{ (e_\ell, r_k, e_j)\ |\ e_i \neq e_\ell \land (e_i, r_k, e_j) \in \calD^+ \} \\
          & \cup
            \{ (e_i, r_k, e_\ell)\ |\ e_j \neq e_\ell \land (e_i, r_k, e_j) \in \calD^+ \}
\end{align*}
To understand the difference between this approach and the CWA (where we assumed
all valid unknown triples were false), let us consider the example in
Figure~\ref{fig:spock-example}. The CWA would generate ``good'' negative triples
such as (\textit{LeonardNimoy, starredIn, StarWars}), (\textit{AlecGuinness,
  {starredIn}, StarTrek}), etc., but also type-consistent but ``irrelevant''
negative triples such as (\textit{BarackObama, {starredIn}, StarTrek}), etc. (We
are assuming (for the sake of this example) there is a type Person but not a
type Actor.) The second approach (based on perturbation) would not generate
negative triples such as (\textit{BarackObama, {starredIn}, StarTrek}), since
\textit{BarackObama} does not participate in any \textit{starredIn} events. This
reduces the size of $\calD^-$, and encourages it to focus on ``plausible''
negatives. (An even better method, used in Section~\ref{sec:kv}, is to generate
the candidate triples from text extraction methods run on the Web. Many of these
triples will be false, due to extraction errors, but they define a good set of
``plausible'' negatives.)

Another option to generate negative examples for training is to make a
\emph{local-closed world assumption} (LCWA)
\citep{galarraga_amie:_2013,dong_knowledge_2014}, in which we assume that a KG
is only \emph{locally} complete. More precisely, if we have observed any triple
for a particular subject-predicate pair $e_i, r_k$, then we will assume that any
non-existing triple $(e_i, r_k, \cdot)$ is indeed false and include them in
$\calD^-$. (The assumption is valid for functional relations, such as
\textit{bornIn}, but not for set-valued relations, such as \textit{starredIn}.)
However, if we have not observed any triple at all for the pair $e_i, r_k$, we
will assume that all triples $(e_i, r_k, \cdot)$ are unknown and \emph{not}
include them in $\calD^-$.

\subsection{Pairwise loss training}
Given that the negative training examples are not always really negative, an
alternative approach to likelihood training is to try to make the probability
(or in general, some scoring function) to be larger for true triples than for
assumed-to-be-false triples. That is, we can define the following objective
function:
\begin{equation}
  \min_{\Theta} \sum_{x^+ \in \calD^+}
  \sum_{x^- \in \calD^-} \loss(f(x^+; \Theta), f(x^-; \Theta))
  + \lambda \operatorname{reg}(\Theta)
\end{equation}
where $\loss(f, f^\prime)$ is a margin-based ranking loss function such as
\begin{equation}
  \loss(f, f') = \max(1 + f^\prime - f, 0) .
\end{equation}
This approach has several advantages. First, it does not assume that negative
examples are necessarily negative, just that they are ``more negative'' than the
positive ones. Second, it allows the $f(\cdot)$ function to be any function, not
just a probability (but we do assume that larger $f$ values mean the triple is
more likely to be correct).

This kind of objective function is easily optimized by \emph{stochastic gradient
  descent} (SGD) \citep{bottou_large-scale_2010}: at each iteration, we just
sample one positive and one negative example. SGD also scales well to large
datasets. However, it can take a long time to converge. On the other hand, as
discussed previously, some models, when combined with the squared loss
objective, can be optimized using alternating least squares (ALS), which is
typically much faster.

\subsection{Model selection}
Almost all models discussed in previous sections include one or more
user-given parameters that are influential for the model's performance (e.g.,
dimensionality of latent feature models, length of relation paths for PRA,
regularization parameter for penalized maximum likelihood training).
Typically, cross-validation over random splits of $\calD$ into training-,
validation-, and test-sets is used to find good values for such parameters
without overfitting (for more information on model selection in machine learning
see e.g.,~\citep{murphy_machine_2012}).
For link prediction and entity resolution,
the \emph{area under the ROC curve} (AUC-ROC) or the \emph{area under the
precision-recall curve} (AUC-PR) are good evaluation criteria.
For data with a large number of negative examples (as it is typically the case
for knowledge graphs), it has been shown that AUC-PR can give a clearer picture of an
algorithm's performance than AUC-ROC~\citep{davis2006relationship}. For entity
resolution, the mean reciprocal rank (MRR) of the correct entity is an
alternative evaluation measure.


\section{Markov random fields}
\label{sec:MRF}
\label{sec:MLN}
\label{sec:mln}

In this section we drop the assumption that the random variables $y_{ijk}$ in
$\tY$ are conditionally independent. However, in the case of relational
data and without the conditional independence assumption, each $y_{ijk}$ can
depend on any of the other $\Nentities \times \Nentities \times \Nreln - 1$
random variables in $\ten{Y}$. Due to this enormous number of possible
dependencies, it becomes quickly intractable to estimate the joint distribution
$\prob(\tY)$ without further constraints, even for very small knowledge graphs.
To reduce the number of potential dependencies and arrive at tractable models,
in this section we develop template-based graphical models that only consider a
small fraction of all possible dependencies.

(See~\cite{KollerBook} for an introduction to graphical models.)

\subsection{Representation}
Graphical models use graphs to encode dependencies between random variables.
Each random variable (in our case, a possible fact $y_{ijk}$) is represented as
a node in the graph, while each dependency between random variables is
represented as an edge. To distinguish such graphs from knowledge graphs, we
will refer to them as \emph{dependency graphs}. It is important to be aware of
their key difference: while knowledge graphs encode the existence of
facts, dependency graphs encode statistical dependencies between random
variables.

To avoid problems with cyclical dependencies, it is common to use
\emph{undirected} graphical models, also called Markov Random Fields
(MRFs).\footnote{Technically, since we are conditioning on some observed
  features $\vx$, this is a Conditional Random Field (CRF), but we will ignore
  this distinction.} A MRF has the following form:
\begin{equation}
  \prob(\tY|\vtheta) = \frac{1}{Z} \prod_{c} \psi(\vy_c|\vtheta)
  \label{eqn:mrf}
\end{equation}
where $\psi(\vy_c |\vtheta) \geq 0$ is a potential function on the $c$-th subset
of variables, in particular the $c$-th clique in the dependency graph, and $Z =
\sum_{\vy} \prod_c \psi(\vy_c|\vtheta)$ is the partition function, which ensures
that the distribution sums to one. The potential functions capture local
correlations between variables in each clique $c$ in the dependency graph. (Note
that in undirected graphical models, the local potentials do not have any
probabilistic interpretation, unlike in directed graphical models.) This
equation again defines a probability distribution over ``possible worlds'',
i.e., over joint distribution assigned to the random variables $\tY$.

The structure of the dependency graph (which defines the cliques in
\cref{eqn:mrf}) is derived from a template mechanism that can be defined in a
number of ways. A common approach is to use \emph{Markov logic}
\citep{richardson_markov_2006}, which is a template language based on logical
formulae:

Given a set of formulae $\Set{F} = \{F_i\}_{i=1}^L$, we create an edge between
nodes in the dependency graph if the corresponding facts occur in at least one
grounded formula. A grounding of a formula $F_i$ is given by the (type
consistent) assignment of entities to the variables in $F_i$. Furthermore, we
define $\psi(\vy_c| \theta)$ such that
\begin{equation}
  \prob(\tY| \vtheta) = \frac{1}{Z} \prod_c \exp(\theta_c x_c)
\end{equation}
where $x_c$ denotes the number of true groundings of $F_c$ in $\tY$, and
$\theta_c$ denotes the weight for formula $F_c$. If $\theta_c > 0$, we prefer
worlds where formula $F_c$ is satisfied; if $\theta_c < 0$, we prefer worlds
where formula $F_c$ is violated. If $\theta_c=0$, then formula $F_c$ is ignored.

\begin{figure*}
  \centering
  \begin{minipage}{.29\textwidth}
    \subfloat[]{\label{fig:mln1}\resizebox{\textwidth}{!}{\begin{tikzpicture}[scale=2.0,baseline,very thick]
    \GraphInit[vstyle=Classic]
    \tikzset{vertex/.style =
      {draw=black,shape=circle,fill=white,minimum size=13pt,circular
        drop shadow}}
    \tikzset{rv/.style =
      {draw=black,shape=diamond,fill=white,minimum size=13pt}}
    \node at (0,2)[vertex,label=above:$a_1$] (a1) {};
    \node at (2,2)[vertex,label=above:$a_2$] (a2) {};
    \node at (0,0)[vertex,label=below:$a_3$] (a3) {};
    \node at (2,0)[vertex,label=below:$c$] (c) {};



    \tikzstyle{EdgeStyle}=[<->,>=stealth,very thick,dashed,draw=red]
    \Edge(a1)(a2);
    \Edge(a1)(a3);
    \Edge(a2)(a3);
    
    \tikzstyle{EdgeStyle}=[->,>=stealth,very thick,dotted,draw=blue]
    \Edge(a1)(c);
    \Edge(a2)(c);
    \Edge(a3)(c);

\end{tikzpicture}

}}
  \end{minipage}
  \hfill
  \begin{minipage}{.33\textwidth}
    \subfloat[]{\label{fig:mln2}\resizebox{\textwidth}{!}{\begin{tikzpicture}[scale=2.0,baseline,very thick]
    \GraphInit[vstyle=Classic]
    \tikzset{vertex/.style =
      {draw=black,shape=circle,fill=white,minimum size=13pt,circular
        drop shadow}}
    \tikzset{rv/.style =
      {draw=black,shape=diamond,fill=white,minimum size=35pt}}
    \node at (0,2)[vertex,label=above:$a_1$] (a1) {};
    \node at (2,2)[vertex,label=above:$a_2$] (a2) {};
    \node at (0,0)[vertex,label=below:$a_3$] (a3) {};
    \node at (2,0)[vertex,label=below:$c$] (c) {};



    \tikzstyle{EdgeStyle}=[<->,>=stealth,very thick,dashed,draw=red]
    \Edge(a1)(a2);
    \Edge(a1)(a3);
    \Edge(a2)(a3);
    
    \tikzstyle{EdgeStyle}=[->,>=stealth,very thick,dotted,draw=blue]
    \Edge(a1)(c);
    \Edge(a2)(c);
    \Edge(a3)(c);

    \node at (0,1)[rv] (m13) {\footnotesize{$m_{13}$}};
    \node at (1,2)[rv] (m12) {\footnotesize{$m_{12}$}};
    \node at (1,1)[rv] (m23) {\footnotesize{$m_{23}$}};
    \node at (1,0)[rv] (p3) {\footnotesize{$p_3c$}};
    \node at (2,1)[rv] (p2) {\footnotesize{$p_2c$}};
    \node at (1.5, 0.5)[rv] (p1) {\footnotesize{$p_1c$}};
\end{tikzpicture}

}}
  \end{minipage}
  \hfill
  \begin{minipage}{.33\textwidth}
    \subfloat[]{\label{fig:mln5}\resizebox{\textwidth}{!}{\begin{tikzpicture}[scale=2.0,baseline,very thick,bend angle=45]
    \GraphInit[vstyle=Classic]
    \tikzset{vertex/.style =
      {draw=black,shape=circle,fill=white,minimum size=13pt,circular
        drop shadow}}
    \tikzset{rv/.style =
      {draw=black,shape=diamond,fill=white,minimum size=35pt}}



    

    \node at (0,1)[rv] (m13) {\footnotesize{$m_{13}$}};
    \node at (1,2)[rv] (m12) {\footnotesize{$m_{12}$}};
    \node at (1,1)[rv] (m23) {\footnotesize{$m_{23}$}};
    \node at (1,0)[rv] (p3) {\footnotesize{$p_3c$}};
    \node at (2,1)[rv] (p2) {\footnotesize{$p_2c$}};
    \node at (1.5, 0.5)[rv] (p1) {\footnotesize{$p_1c$}};
    
    \tikzstyle{EdgeStyle}=[-,>=stealth,very thick]
    \Edge(m13)(p1); \Edge(m13)(p3);
    \Edge(m12)(p1); \Edge(m12)(p2);
    \Edge(m23)(p3); \Edge(m23)(p2);
    \Edge(p1)(p3); \Edge(p1)(p2);
    \Edge[style=bend right](p3)(p2);
\end{tikzpicture}

}}
  \end{minipage}
  \caption{
    \protect\subref{fig:mln1} A small KG.
    There are 4 entities (circles): 3 adults ($a_1$, $a_2$, $a_3$) and 1 child c
    There are 2 types of edges: adults may or may
    not be married to each other, as indicated by the red dashed edges,
    and  the adults may or may not be parents of the child, as indicated by the
    blue dotted edges.
    \protect\subref{fig:mln2} We add binary random variables (represented by
    diamonds) to each KG edge.
    \protect\subref{fig:mln5} We drop the entity nodes, and add edges between
    the random variables that belong to the same clique potential, resulting in
    a standard MRF.}
\end{figure*}
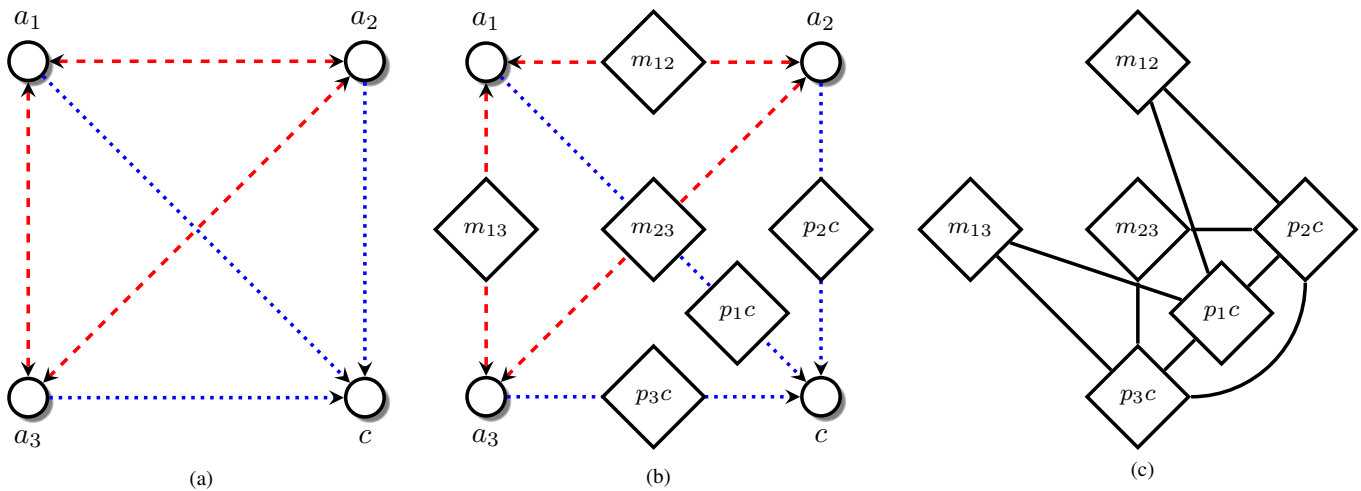

To explain this further, consider a KG involving two types of entities,
adults and children, and two types of relations, \emph{parentOf} and
\emph{marriedTo}. Figure~\ref{fig:mln1} depicts a sample KG with three adults and one child.
Obviously, these relations (edges) are correlated, since
people who share a common child are often married, while people rarely marry their own children.
In Markov logic, we represent these dependencies using formulae such as:
\begin{align*}
F_1 & : (x, \textit{parentOf}, z) \land (y, \textit{parentOf}, z) \Rightarrow (x, \textit{marriedTo}, y)\\
F_2 & : (x, \textit{marriedTo}, y) \Rightarrow \lnot (y, \textit{parentOf}, x)
\end{align*}
Rather than encoding the rule that adults cannot marry their own children using
a formula, we will encode this as a hard constraint into the type system.
Similarly, we only allow adults to be parents of children. Thus, there are 6
possible facts in the knowledge graph. To create a dependency graph for this KG
and for this set of logical formulae $\Set{F}$, we assign a binary random
variable to each possible fact, represented by a diamond in
Figure~\ref{fig:mln2}, and create edges between these nodes if the corresponding
facts occur in grounded formulae $F_1$ or $F_2$. For instance, grounding $F_1$
with $x = a_1$, $y = a_3$, and $z = c$, creates the edges $m_{13} \to p_1c$,
$m_{13} \to p_3c$, and $p_1c \to p_3c$. The full dependency graph is shown in
\cref{fig:mln5}.

The process of generating the MRF graph by applying templated rules to a set of
entities is known as {\em grounding} or {\em instantiation}. We note that the
topology of the resulting graph is quite different from the original KG. In
particular, we have one node per possible KG edge, and these nodes are densely
connected. This can cause computational difficulties, as we discuss below.

\subsection{Inference}
The inference problem consists of estimating the most probable configuration,
$\vy^* = \arg \max_{\vy} p(\vy|\vtheta)$, or the posterior marginals
$p(y_i|\vtheta)$. In general, both of these problems are computationally
intractable \cite{KollerBook}, so heuristic approximations must be used.

One approach for computing posterior marginals is to use Gibbs sampling (see,
or example, \cite{niu_elementary:_2012,zhang_towards_2013}) or MC-SAT
\cite{Poon2006}. One approach for computing the MAP estimate is to use the MPLP
(max product linear programming) method \cite{Globerson2007}. See
\cite{KollerBook} for more details.

If one restricts the class of potential functions to be just disjunctions (using
OR and NOT, but no AND), then one obtains a (special case of) hinge loss MRF
(HL-MRFs) \cite{bach:arxiv15}, for which efficient convex algorithms can be
applied, based on a continuous relaxation of the binary random variables.
Probabilistic Soft Logic (PSL)~\citep{kimmig_short_2012} provides a convenient
form of ``syntactic sugar'' for defining HL-MRFs, just as MLNs provide a form of
syntactic sugar for regular (boolean) MRFs. HL-MRFs have been shown to scale to
fairly large knowledge bases \citep{Pujara2015}.

\subsection{Learning}
The ``learning'' problem for MRFs deals with specifying the form of the
potential functions (sometimes called ``structure learning'') as well as the
values for the numerical parameters $\vtheta$. In the case of MRFs for KGs, the
potential functions are often specified in the form of logical rules, as
illustrated above. In this case, structure learning is equivalent to rule
learning, which has been studied in a number of published works (see
Section~\ref{sec:PRArules} and \cite{galarraga_amie:_2013,yang_embedding_2014}).

The parameter estimation problem (which is usually cast as maximum likelihood or
MAP estimation), although convex, is in general quite expensive, since it needs
to call inference as a subroutine. Therefore, various faster approximations,
such as pseudo likelihood, have been developed (cf.\ relational dependency
networks \cite{neville_relational_2007}).

\subsection{Discussion}
Although approaches based on MRFs are very flexible, it is in general harder to
make scalable inference and devise learning algorithms for this model class,
compared to methods based on observable or even latent feature models. In this
article, we have chosen to focus primarily on latent and graph feature models
because we have more experience with such methods in the context of KGs.
However, all three kinds of approaches to KG modeling are useful.


\section{Knowledge Vault: relational learning for knowledge base construction}
\label{sec:exper}
\label{sec:kv}
\label{sec:KV}

\newcommand{\ntriples}{1.6B\xspace} 
\newcommand{\ntriplesseven}{324M\xspace}
\newcommand{\ntriplesnine}{271M\xspace}
\newcommand{\nentities}{45M\xspace}  
\newcommand{\npredicates}{4469\xspace} 
\newcommand{\nclasses}{1100\xspace} 

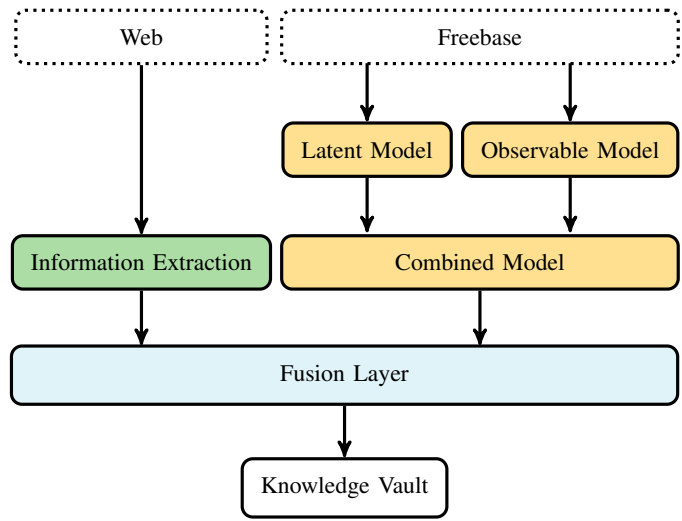
\begin{figure}
  \begin{tikzpicture}
\tikzset{
>=stealth',
  vertex/.style={
    rectangle, 
    rounded corners, 
    font=\small,
    inner sep=0.7em,
    draw=black, very thick,
    text centered}};
\node at (-2.7,4.5)[vertex,minimum width=9.5em,dotted] (web) {Web};
\node at (1.8,4.5)[vertex,minimum width=15em,dotted] (fb) {Freebase};
\node at (0.3,3)[vertex,fill=ccls3] (mlp) {Latent Model};
\node at (3,3)[vertex,fill=ccls3] (pra) {Observable Model};
\node at (1.8,1.5)[vertex,minimum width=15em,fill=ccls3] (com) {Combined Model};
\node at (-2.7,1.5)[vertex,fill=ccls43] (txt) {Information Extraction};
\node at (0,0)[vertex,minimum width=\columnwidth,fill=ccls4] (fus) {Fusion Layer};
\node at (0,-1.5)[vertex] (kv) {Knowledge Vault};

\draw (web.south) edge[->,very thick] (txt.north);
\draw (0.3,4.15) edge[->,very thick] (mlp.north);
\draw (3,4.15) edge[->,very thick] (pra.north);
\draw (mlp.south) edge[->,very thick] (0.3,1.9);
\draw (pra.south) edge[->,very thick] (3,1.9);
\draw (com.south) edge[->,very thick] (1.8,0.4);
\draw (txt.south) edge[->,very thick] (-2.7,0.4);
\draw (fus.south) edge[->,very thick] (kv.north);
\end{tikzpicture}

  \centering
  \caption{Architecture of the Knowledge Vault.}
  \label{fig:kv-arch}
\end{figure}

The Knowledge Vault (KV)~\cite{dong_knowledge_2014} is a very large-scale
automatically constructed knowledge base, which follows the Freebase schema (KV
uses the 4469 most common predicates). It is constructed in three steps. In the
first step, facts are extracted from a host of Web sources such as natural
language text, tabular data, page structure, and human annotations (the
extractors are described in detail in~\cite{dong_knowledge_2014}). Second, an
SRL model is trained on Freebase to serve as a ``prior'' for computing the
probability of (new) edges. Finally, the confidence in the automatically
extracted facts is evaluated using both the extraction scores and the prior SRL
model.

The Knowledge Vault uses a combination of latent and observable models to
predict links in a knowledge graph. In particular, it employs the ER-MLP model
(\cref{sec:MLP}) as a latent feature model and PRA (\cref{sec:pra}) as a graph
feature model. In order to combine the two models, KV uses stacking
(\cref{sec:other-combining}). To evaluate the link prediction performance, these
models were applied to a subset of Freebase. The ER-MLP system achieved an area
under the ROC curve (AUC-ROC) of 0.882, and the PRA approach achieved an almost
identical AUC-ROC of 0.884. The combination of both methods further increased
the AUC-ROC to 0.911. To predict the final score of a triple, the scores from
the combined link-prediction model are further combined with various features
derived from the extracted triples. These include, for instance, the confidence
of the extractors and the number of (de-duplicated) Web pages from which the
triples were extracted. \cref{fig:kv-arch} provides a high level overview of the
Knowledge Vault architecture.

Let us give a qualitative example of the benefits of combining the prior with
the extractors (i.e., the Fusion Layer in \cref{fig:kv-arch}). Consider an
extracted triple corresponding to the following relation:\footnote{%
  For clarity of presentation we show a simplified triple. Please
  see~\citep{dong_knowledge_2014} for the actually extracted triples including
  compound value types (CVT). }
\begin{center}
  \textit{(Barry Richter, attended, University of Wisconsin-Madison)}.
\end{center}
The extraction confidence for this triple (obtained by fusing multiple
extraction techniques) is just 0.14, since it was based on the following two
rather indirect statements:\footnote{%
  Source:
  \url{http://www.legendsofhockey.net/LegendsOfHockey/jsp/SearchPlayer.jsp?player=11377}
}
\begin{quote}
  \itshape
  In the fall of 1989, Richter accepted a scholarship
  to the University of Wisconsin, where he played for
  four years and earned numerous individual accolades \ldots
\end{quote}

and\footnote{Source: \url{http://host.madison.com/sports/high-school/hockey/numbers-dwindling-for-once-
mighty-madison-high-school-hockey-programs/article_95843e00-ec34-11df-9da9-001cc4c002e0.html}}

\begin{quote}
  \itshape
  The Polar Caps' cause has been helped by the impact of
  knowledgable coaches such as Andringa, Byce and former
  UW teammates Chris Tancill and Barry Richter.
\end{quote}
However, we know from Freebase that Barry Richter was born and raised in
Madison, Wisconsin. According to the prior model, people who were born and
raised in a particular city often tend to study in the same city. This increases
our prior belief that Richter went to school there, resulting in a final fused
belief of 0.61.

Combining the prior model (learned using SRL methods) with the information
extraction model improved performance significantly, increasing the number of
high confidence triples\footnote{Triples with the calibrated probability of
  correctness above 90\%.} from 100M (based on extractors alone) to 271M (based
on extractors plus prior). The Knowledge Vault is one of the largest
applications of SRL to knowledge base construction to date. See
\cite{dong_knowledge_2014} for further details.


\section{Extensions and Future Work}
\label{sec:extension}

\subsection{Non-binary relations}
\label{sec:higherOrderRelations}

So far we completely focussed on binary relations; here we discuss how
relations of other cardinalities can be handled.

\subsubsection*{Unary relations}
Unary relations refer to statements on properties of entities, e.g., the height
of a person. Such data can naturally be represented by a matrix, in which rows
represent entities, and columns represent attributes.
\citet{nickel_factorizing_2012} proposed a joint tensor-matrix factorization
approach to learn simultaneously from binary and unary relations via a shared
latent representation of entities. In this case, we may also need to modify the
likelihood function, so it is Bernoulli for binary edge variables, and Gaussian
(say) for numeric features and Poisson for count data
(see~\cite{krompas_probabilistic_2014}).

\subsubsection*{Higher-arity relations}
In knowledge graphs, higher-arity relations are typically expressed via multiple
binary relations. In \cref{sec:kg}, we expressed the ternary relationship
\mbox{\textit{playedCharacterIn(LeonardNimoy, Spock, StarTrek-1)}} via two
binary relationships \textit{(LeonardNimoy, played, Spock)} and \textit{(Spock,
  characterIn, StarTrek-1)}. However, there are multiple actors who played Spock
in different Star Trek movies, so we have lost the correspondence between
Leonard Nimoy and StarTrek-1. To model this using binary relations without loss
of information, we can use auxiliary nodes to identify the respective
relationship. For instance, to model the relationship
\mbox{\textit{playedCharacterIn(LeonardNimoy, Spock, StarTrek-1)}}, we can write
\begin{triplelist}
  (LeonardNimoy, & actor, &  \underline{MovieRole-1})\\
  (\underline{MovieRole-1}, & movie, &  StarTreck-1)\\
  (\underline{MovieRole-1}, & character, &  Spock)
\end{triplelist}
where we used the auxiliary entity \underline{\textit{MovieRole-1}} to uniquely
identify this particular relationship. In most applications auxiliary entities
get an identifier; if not they are referred to as \emph{blank nodes}. In
Freebase auxiliary nodes are called \emph{Compound Value Types} (CVT).

Since higher-arity relations involving time and location are relatively common,
the YAGO2 project extended the SPO triple format to the \textit{(subject,
  predicate, object, time, location)} (SPOTL) format to model temporal and
spatial information about relationships explicitly, without transforming them to
binary relations~\cite{hoffart_yago2:_2013}. Furthermore, there has also been
work on extracting higher-arity relations directly from natural
language~\citep{li2015improvement}.

A related issue is that the truth-value of a fact can change over time. For
example, Google's current CEO is Larry Page, but from 2001 to 2011 it was Eric
Schmidt. Both facts are correct, but only during the specified time interval.
For this reason, Freebase allows some facts to be annotated with beginning and
end dates, using CVT (compound value type) constructs, which represent n-ary
relations via auxiliary nodes. In the future, it is planned to extend the KV
system to model such temporal facts. However, this is non-trivial, since it is
not always easy to infer the duration of a fact from text, since it is not
necessarily related to the timestamp of the corresponding source (cf.
\citep{ji_tackling_2013}).

As an alternative to the usage of auxiliary nodes, a set of $n-$th-arity
relations can be represented by a single $n+1-$th-order tensor. RESCAL can
easily be generalized to higher-arity relations and can be solved by
higher-order tensor factorization or by neural network models with the
corresponding number of entity representations as
inputs~\citep{krompas_probabilistic_2014}.

\subsection{Hard constraints: types, functional constraints, and others}
Imposing hard constraints on the allowed triples in knowledge graphs can be
useful. Powerful ontology languages such as the Web Ontology Language
(OWL)~\citep{mcguinness_owl_2004} have been developed, in which complex
constraints can be formulated. However, reasoning with ontologies is
computationally demanding, and hard constraints are often violated in real-world
data~\citep{hogan_weaving_2010,halpin_when_2010}. Fortunately, machine learning
methods can be robust in the face of contradictory evidence.

\subsubsection*{Deterministic dependencies}
Triples in relations such as \textit{subClassOf} and \textit{isLocatedIn} follow
clear deterministic dependencies such as transitivity. For example, if Leonard
Nimoy was born in Boston, we can conclude that he was born in Massachusetts,
that he was born in the USA, that he was born in North America, etc. One way to
consider such ontological constraints is to precompute all true triples that can
be derived from the constraints and to add them to the knowledge graph prior to
learning. The precomputation of triples according to ontological constraints is
also called \emph{materialization}. However, on large knowledge graphs, full
materialization can be computationally demanding.

\subsubsection*{Type constraints}
Often relations only make sense when applied to entities of the right type. For
example, the domain and the range of \emph{marriedTo} is limited to entities
which are persons. Modelling type constraints explicitly requires complex manual
work. An alternative is to learn approximate type constraints by simply
considering the observed types of subjects and objects in a relation. The
standard RESCAL model has been extended by~\citet{krompas_large-scale_2014} and
\citet{chang_typed_2014} to handle type constraints of relations efficiently. As
a result, the rank required for a good RESCAL model can be greatly reduced.
Furthermore, \citet{riedel_relation_2013} considered learning latent
representations for the \emph{argument slots} in a relation to learn the correct
types from data.

\subsubsection*{Functional constraints and mutual exclusiveness}
Although the methods discussed in \cref{sec:lfm,sec:observable} can model
long-range and global dependencies between triples, they do not explicitly
enforce functional constraints that induce mutual exclusivity between possible
values. For instance, a person is born in exactly one city, etc. If one of the
these values is observed, then observable graph models can prevent other values
from being asserted, but if all the values are unknown, the resulting mutual
exclusion constraint can be hard to deal with computationally.

\subsection{Generalizing to new entities and relations}
\label{sec:newEntities}

In addition to missing facts, there are many entities that are mentioned on the
Web but are currently missing in knowledge graphs like Freebase and YAGO. If new
entities or predicates are added to a KG, one might want to avoid retraining the
model due to runtime considerations. Given the current model and a set of newly
observed relationships, latent representations of new entities can be calculated
approximately in both tensor factorization models and in neural networks, by
finding representations that explain the newly observed relationships relative
to the current model. Similarly, it has been shown that the relation-specific
weights $\mW_k$ in the RESCAL model can be calculated efficiently for new
relation types given already derived latent representations of
entities~\citep{krompas_querying_2014}.

\subsection{Querying probabilistic knowledge graphs}
RESCAL and KV can be viewed as probabilistic databases (see, e.g.,
\citep{suciu_probabilistic_2011,wang_bayesstore:_2008}). In the Knowledge Vault,
only the probabilities of triples are queried. Some applications might require
more complex queries such as: \textit{Who is born in Rome and likes someone who
  is a child of Albert Einstein}. It is known that queries involving joins
(existentially quantified variables) are expensive to calculate in probabilistic
databases (\citep{suciu_probabilistic_2011}). In~\citet{krompas_querying_2014},
it was shown how some queries involving joins can be efficiently handled within
the RESCAL framework.

\subsection{Trustworthiness of knowledge graphs}
\label{sec:trust}
Automatically constructed knowledge bases are only as good as the sources from
which the facts are extracted. Prior studies in the field of data fusion have
developed numerous approaches for modelling the correctness of information
supplied by multiple sources in the presence of possible data conflicts (see
\citep{Bleiholder:2009:DF,Li:2012:TFD} for recent surveys). However, the key
assumption in data fusion---namely, that the facts provided by the sources are
indeed stated by them---is often violated when the information is extracted
automatically. If a given source contains a mistake, it could be because the
source actually contains a false fact, or because the fact has been extracted
incorrectly. A recent study \citep{Dong:2014:DFK} has formulated the problem of
knowledge fusion, where the above assumption is no longer made, and the
correctness of information extractors is modeled explicitly. A follow-up study
by the authors \citep{Dong:2015:KTE} developed several approaches for solving
the knowledge fusion problem, and applied them to estimate the trustworthiness
of facts in the Knowledge Vault (cf.\ Section~\ref{sec:kv}).


\section{Concluding Remarks}
\label{sec:concl}

Knowledge graphs have found important applications in question answering,
structured search, exploratory search, and digital assistants. We provided a
review of state-of-the-art statistical relational learning (SRL) methods applied
to very large knowledge graphs. We also demonstrated how statistical relational
learning can be used in conjunction with machine reading and information
extraction methods to automatically build such knowledge repositories. As a
result, we showed how to create a truly massive, machine-interpretable
``semantic memory'' of facts, which is already empowering numerous practical
applications. However, although these KGs are impressive in their size, they
still fall short of representing many kinds of knowledge that humans possess.
Notably missing are representations of ``common sense'' facts (such as the fact
that water is wet, and wet things can be slippery), as well as ``procedural'' or
how-to knowledge (such as how to drive a car or how to send an email).
Representing, learning, and reasoning with these kinds of knowledge remains the
next frontier for AI and machine learning.


\section{Appendix}
\label{sec:appendix}

\subsection{RESCAL is a special case of NTN}
\label{sec:rescalAsNTN}

Here we show how the
RESCAL model of Section~\ref{sec:rescal}
is a special case of the
 neural tensor model (NTN) of Section~\ref{sec:otherNN}.
To see this, note that RESCAL has the form
\begin{align}
f_{ijk}^{\text{RESCAL}} & = \ve_i^\transp \mW_k \ve_j
 = \vw_k^\transp [\ve_j \kron \ve_i]
\label{eqn:rescal}
\end{align}
Next, note that
\[
\vv \kron \vu = \vect{\vu \vv^\transp}
= [\vu^\transp \mB^1 \vv,  \ldots, \vu^\transp \mB^{n} \vv]
\]
where $n = |u| |v|$, and $\mB^k$ is a matrix of all 0s except for a single 1 element in the $k$'th position, which ``plucks out'' the corresponding  
entries from the $\vu$ and $\vv$ matrices.
For example, 
\begin{eqnarray}
\begin{pmatrix} u_1 \\ u_2 \end{pmatrix}
\begin{pmatrix} v_1 & v_2 \end{pmatrix}
 &=&
\left[
\begin{pmatrix} u_1 \\ u_2 \end{pmatrix}^\transp
\begin{pmatrix} 1 & 0 \\ 0 & 0 \end{pmatrix}
\begin{pmatrix} v_1 \\ v_2 \end{pmatrix}, \right.
\nonumber \\
&& \left.
\ldots, \;
\begin{pmatrix} u_1 \\ u_2 \end{pmatrix}^\transp
\begin{pmatrix} 0 & 0 \\ 0 & 1 \end{pmatrix}
\begin{pmatrix} v_1 \\ v_2 \end{pmatrix}
\right].
\end{eqnarray}
In general, define $\delta_{ij}$ as a matrix of all 0s except
for entry $(i,j)$ which is 1.
Then if we define
$\tB_k = [\delta_{1,1}, \ldots, \delta_{H_e,H_e}]$
we have
\[
\vh_{ijk}^b = 
\left[\ve_i^\transp \mB_k^1 \ve_j ,
\ldots,\ve_i^\transp \mB_k^{H_b} \ve_j \right]
 = \ve_j \kron \ve_i
\]
Finally, if we define $\mA_k$ as the empty matrix (so $h_{ijk}^a$ is undefined),
and $g(u)=u$ as the identity function, then the NTN equation
\[
f_{ijk}^{\text{NTN}} = \vw_k^\transp g([\vh_{ijk}^a; \vh_{ijk}^b])
\]
matches Equation~\ref{eqn:rescal}.

\section*{Acknowledgment}

Maximilian Nickel acknowledges support by the Center for Brains, Minds and
Machines (CBMM), funded by NSF STC award CCF-1231216. 
Volker Tresp acknowledges support by the German Federal Ministry for Economic
Affairs and Energy, technology program ``Smart Data'' (grant 01MT14001).

\ifCLASSOPTIONcaptionsoff
  \newpage
\fi



\bibliographystyle{IEEEtran-max}
\bibliography{IEEE-proc-bibtex}

%
%
%

%

\begin{IEEEbiography}[{\includegraphics[width=1in,height=1.25in,clip,keepaspectratio]{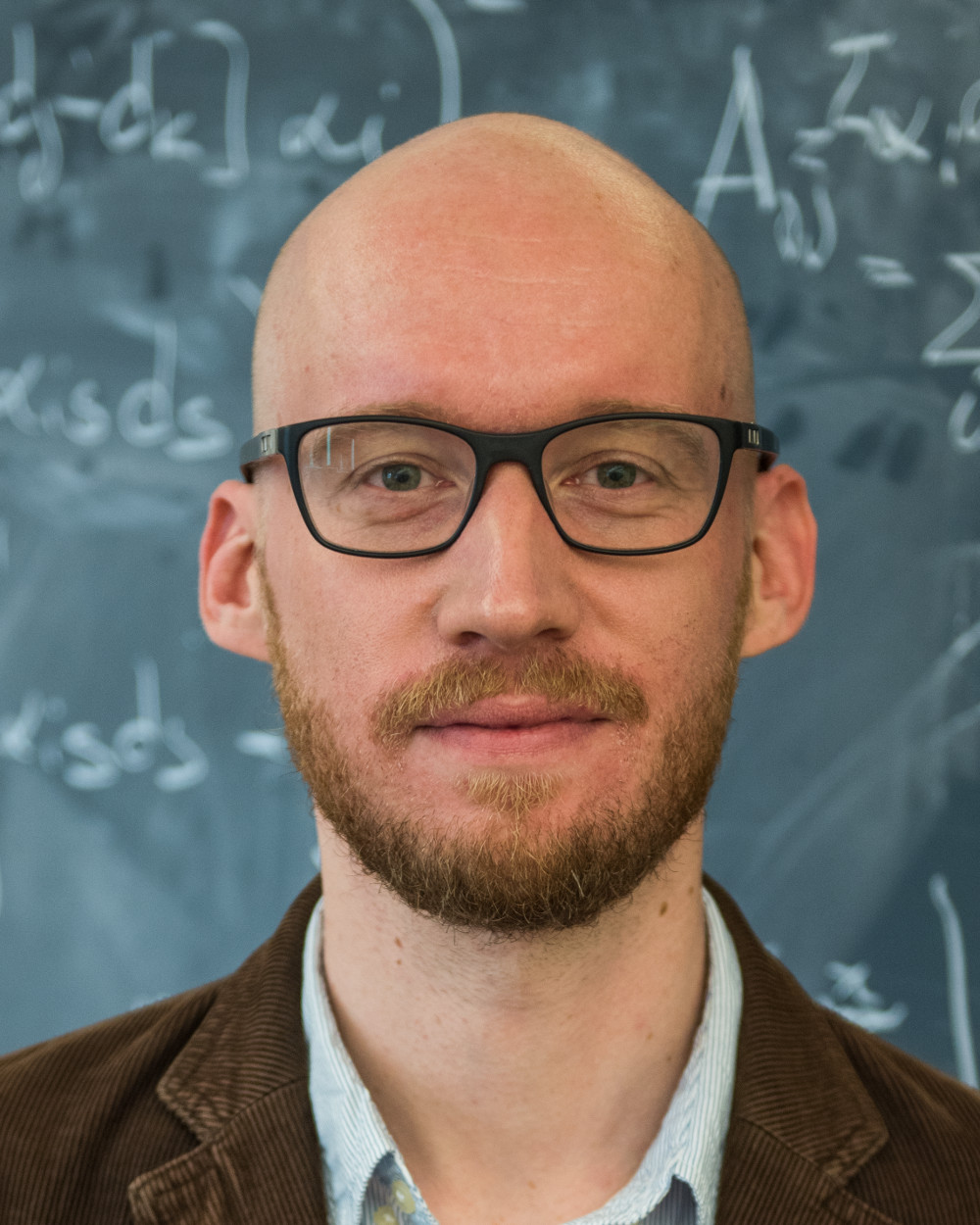}}]{Maximilian Nickel}
  is a postdoctoral fellow with the Laboratory for Computational and Statistical
  Learning at MIT and the Istituto Italiano di Tecnologia. Furthermore, he is with the
  Center for Brains, Minds, and Machines at MIT. In 2013, he received his PhD
  with summa cum laude from the Ludwig Maximilian University in Munich. From 2010
  to 2013 he worked as a research assistant at Siemens Corporate Technology, Munich. His
  research centers around machine learning from relational knowledge
  representations and graph-structured data as well as its applications in artificial
  intelligence and cognitive science.
\end{IEEEbiography}

\begin{IEEEbiography}[{\includegraphics[width=1in,height=1.25in,clip,keepaspectratio]{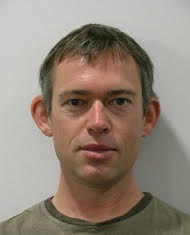}}]{Kevin Murphy}
  is a research scientist at Google in Mountain View, California,
  where he works on AI, machine learning, computer vision, knowledge base
  construction and natural language processing. Before joining Google in 2011,
  he was an associate professor of computer science and statistics at the
  University of British Columbia in Vancouver, Canada. Before starting at UBC in
  2004, he was a postdoc at MIT. Kevin got his BA from U. Cambridge, his MEng
  from U. Pennsylvania, and his PhD from UC Berkeley. He has published over 80
  papers in refereed conferences and journals, as well as an 1100-page textbook
  called ``Machine Learning: a Probabilistic Perspective'' (MIT Press, 2012),
  which was awarded the 2013 DeGroot Prize for best book in the field of
  Statistical Science. Kevin is also the (co) Editor-in-Chief of JMLR (the
  Journal of Machine Learning Research).
\end{IEEEbiography}

\begin{IEEEbiography}[{\includegraphics[width=1in,height=1.25in,clip,keepaspectratio,trim=20
    20 20 0]{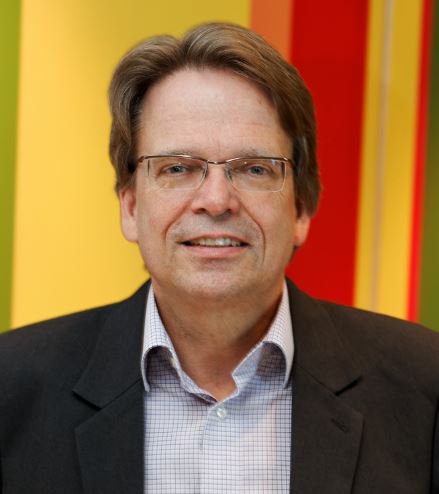}}]{Volker Tresp}
  received a Diploma degree from the University of Goettingen,
  Germany, in 1984 and the M.Sc. and Ph.D. degrees from Yale University, New
  Haven, CT, in 1986 and 1989 respectively. Since 1989 he is the head of various
  research teams in machine learning at Siemens, Research and Technology. He
  filed more than 70 patent applications and was inventor of the year of Siemens
  in 1996. He has published more than 100 scientific articles and administered
  over 20 Ph.D. theses. The company Panoratio is a spin-off out of his team. His
  research focus in recent years has been „Machine Learning in Information
  Networks“ for modelling Knowledge Graphs, medical decision processes and
  sensor networks. He is the coordinator of one of the first nationally funded
  Big Data projects for the realization of „Precision Medicine“. In 2011 he
  became a Honorarprofessor at the Ludwig Maximilian University of Munich where
  he teaches an annual course on Machine Learning.
\end{IEEEbiography}

\begin{IEEEbiography}[{\includegraphics[width=1in,height=1.25in,clip,keepaspectratio,trim=0
    70 0 0]{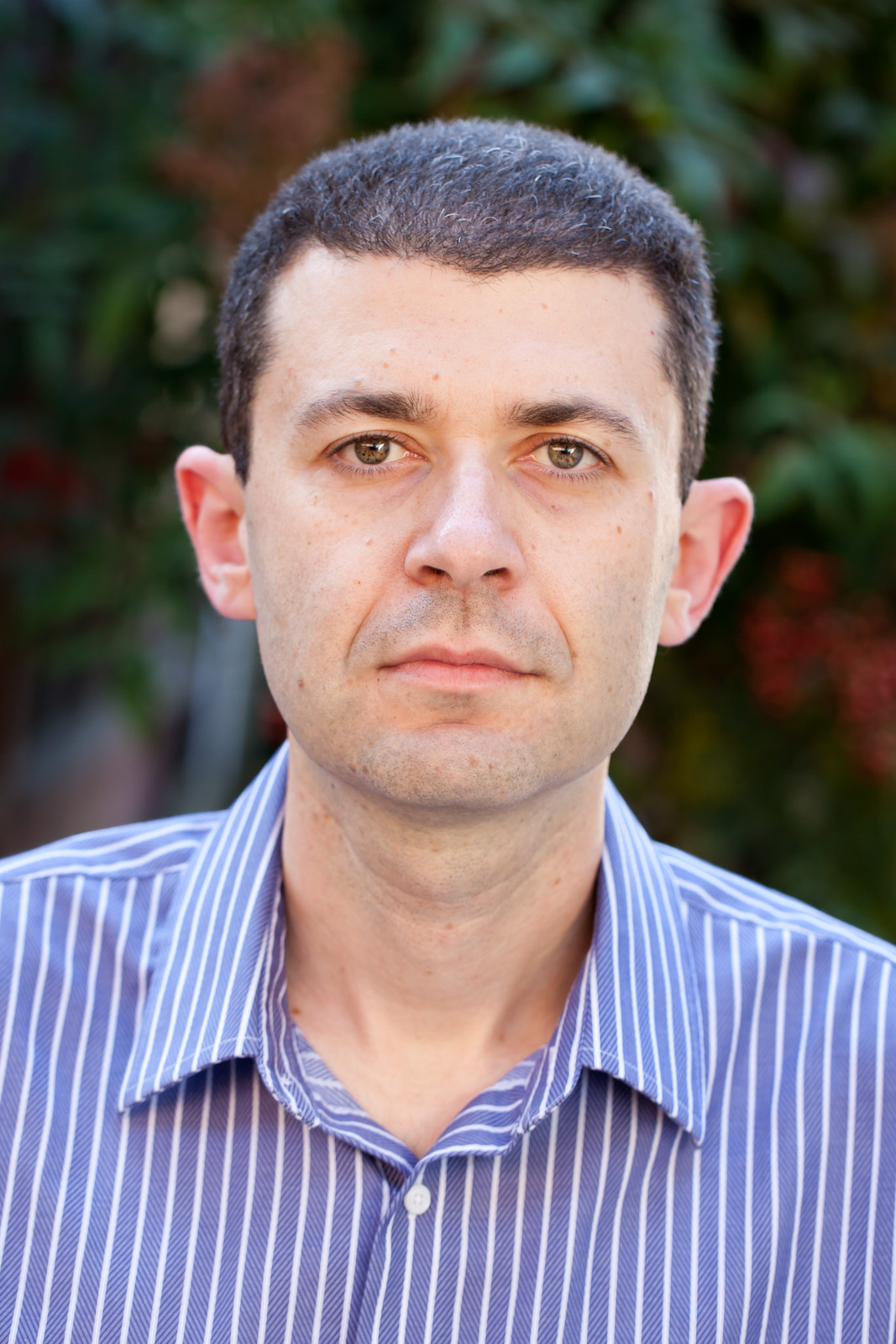}}]{Evgeniy Gabrilovich}
  is a senior staff research scientist at Google, where
  he works on knowledge discovery from the web. Prior to joining Google in 2012,
  he was a director of research and head of the natural language processing and
  information retrieval group at Yahoo! Research. Evgeniy is an ACM
  Distinguished Scientist, and is a recipient of the 2014 IJCAI-JAIR Best Paper
  Prize. He is also a recipient of the 2010 Karen Sparck Jones Award for his
  contributions to natural language processing and information retrieval. He
  earned his PhD in computer science from the Technion - Israel Institute of
  Technology.
\end{IEEEbiography}







\end{document}